\documentclass[12pt]{article}
\usepackage{titling}

\usepackage{pdfpages}

\usepackage{titlesec}
\titleformat{\section}
       {\normalfont\fontsize{14}{17}\bfseries}{\thesection}{1em}{}
\titleformat{\subsection}
       {\normalfont\fontsize{12}{17}\bfseries\itshape}{\thesubsection}{1em}{}
\titleformat{\subsubsection}
       {\normalfont\fontsize{12}{17}\itshape}{\thesubsubsection}{1em}{}

\usepackage{newtxtext} 

\usepackage[authordate, backend=bibtex, uniquename=false, mincitenames=1, maxcitenames=2, citetracker]{biblatex-chicago} 

\usepackage[a4paper, margin=2.5cm]{geometry}
\usepackage{authblk}

\usepackage{setspace}

\usepackage[hidelinks]{hyperref} 
\usepackage{orcidlink} 
\usepackage[french,english]{babel} 
\usepackage[utf8]{inputenc}	
\usepackage{url} 
\usepackage{siunitx} 
\usepackage{float} 
\usepackage[]{graphicx} 

\usepackage{subcaption} 
\usepackage{algorithm}
\usepackage[noend]{algpseudocode}

\usepackage{xcolor}
\usepackage{tabularx} 

\usepackage{amsmath}  
\usepackage{amssymb} 
\usepackage{amsbsy} 

\newcommand{\samplefeature}[0]{x} 
\newcommand{\sample}[0]{\mathbf{\samplefeature}} 
\newcommand{\samplemat}[0]{\mathbf{\MakeUppercase{\samplefeature}}} 
\newcommand{\sampledim}[0]{c} 
\newcommand{\classlabel}[0]{y} 

\newcommand{\preevent}[0]{u} 
\newcommand{\postevent}[0]{v} 
\newcommand{\preimage}[0]{\mathbf{\MakeUppercase{\preevent}} } 
\newcommand{\postimage}[0]{\mathbf{\MakeUppercase{\postevent}} } 
\newcommand{\prevec}[0]{\mathbf{\preevent} } 
\newcommand{\postvec}[0]{\mathbf{\postevent} } 
\newcommand{\hatpreimage}[0]{\hat{\preimage} } 
\newcommand{\hatpostimage}[0]{\hat{\postimage} } 
\newcommand{\hatprevec}[0]{\mathbf{\hat{\preevent}} } 
\newcommand{\hatpostvec}[0]{\mathbf{\hat{\postevent}} } 
\newcommand{\nchanpre}[0]{\sampledim_\preevent } 
\newcommand{\nchanpost}[0]{\sampledim_\postevent } 
\newcommand{\imh}[0]{h} 
\newcommand{\imw}[0]{w} 
\newcommand{\imsize}[0]{\imh \times \imw} 

\newcommand{\codespace}[0]{Z} 
\newcommand{\preencoder}[0]{E_\preevent}   
\newcommand{\postencoder}[0]{E_\postevent} 
\newcommand{\predecoder}[0]{D_\preevent}   
\newcommand{\postdecoder}[0]{D_\postevent} 
\newcommand{\encodepre}[1]{\preencoder(#1)}   
\newcommand{\encodepost}[1]{\postencoder(#1)} 
\newcommand{\decodepre}[1]{\predecoder(#1)}   
\newcommand{\decodepost}[1]{\postdecoder(#1)} 
\newcommand{\codepre}[0]{\codespace_\preevent}   
\newcommand{\codepost}[0]{\codespace_\postevent} 

\newcommand{\prob}[0]{P} 
\newcommand{\pdf}[0]{p} 
\newcommand{\pripos}[0]{\prob(\classlabel = 1)} 
\newcommand{\prineg}[0]{\prob(\classlabel = 0)} 
\newcommand{\posiset}[0]{\mathbb{P}} 
\newcommand{\unlaset}[0]{\mathbb{U}} 
\newcommand{\negaset}[0]{\mathbb{N}} 
\newcommand{\dataset}[0]{\mathbb{\MakeUppercase{\samplefeature}}} 
\newcommand{\npos}[0]{|\posiset|} 
\newcommand{\latentvar}[0]{z} 
\newcommand{\responsibility}[1]{\gamma(\latentvar_{#1})} 

\newcommand{\fte}[0]{forest-tundra ecotone} 

\title{Towards Targeted Change Detection with Heterogeneous Remote Sensing Images for Forest Mortality Mapping }

\author[a]{Jørgen~A.~Agersborg \orcidlink{0000-0002-1526-6659}
}
\author[a]{Luigi~T.~Luppino \orcidlink{0000-0002-1288-817X}}
\author[b, a]{Stian~Normann~Anfinsen \orcidlink{0000-0002-3758-4295}}
\author[c]{Jane~Uhd~Jepsen \orcidlink{0000-0003-1517-1569}}

\affil[a]{Department of Physics and Technology, UiT The Arctic University of Norway, Tromsø, Norway}
\affil[b]{NORCE Norwegian Research Centre, Tromsø, Norway}
\affil[c]{Norwegian Institute for Nature Research, Tromsø, Norway}
\date{}

\begin{document}

\pretitle{\begin{center}\large}
\posttitle{\par\end{center}\vspace{0.5cm}}
\preauthor{\normalfont\normalsize\begin{center}\begin{tabular}[t]{c}}
\postauthor{\end{tabular}\end{center}\vspace{0.5cm}}

\maketitle

\doublespacing

\begin{abstract}
\label{sec:abstract}
Several generic methods have recently been developed
for change detection in heterogeneous remote sensing data, such as images from synthetic aperture radar (SAR) and multispectral radiometers.
However, these are not well suited to detect weak signatures of certain disturbances of ecological systems.
To resolve this problem we propose a new approach based on image-to-image translation and one-class classification (OCC).
We aim to map forest mortality caused by an outbreak of geometrid moths in a sparsely forested forest-tundra ecotone using multisource satellite images.
The images preceding and following the event are collected by Landsat-5 and RADARSAT-2, respectively.
Using a recent deep learning method for change-aware image translation, we compute difference images in both satellites' respective domains.
These differences are stacked with the original pre- and post-event images and passed to an OCC trained on a small sample from the targeted change class.
The classifier produces a credible map of the complex pattern of forest mortality.
\end{abstract}

\selectlanguage{english}

\section{Introduction}
\label{sec:intro}
The \fte{}, the sparsely forested transition zone between northern-boreal forest and low arctic tundra, is changing rapidly with a warming climate \parencite{aba2013caff}. In particular, changes in the distribution of woody vegetation cover through shrub encroachment, tree line advance, and altered pressure from browsers and forest pests, modify the structural and functional attributes of the \fte{} with implications for biodiversity and regional climate feedbacks.

In Northern Norway, mountain birch (\emph{Betula pubescens} var.\ \emph{pumila}) forms the treeline ecotone towards the treeless Low Arctic tundra. In this region, periodic outbreaks by forest defoliators is the most important natural disturbance factor and the only cause of large scale forest die-off. In recent decades, such pest outbreaks have been intensified due to range expansions thought to be linked to more benign climatic conditions \parencite{jepsen2008moth,jepsen2011rapid}. Today, two species of geometrid moths; the autumnal moth (\emph{Epirrita autumnata}) and the winter moth (\emph{Operophtera brumata}) have overlapping population outbreaks at approximately decadal intervals, some of which cause regional scale defoliation and tree and shrub mortality in the forest-tundra ecotone. Outbreaks can thus lead to a reduction in forested areas as well as cascading effects on other species \parencite{biuw2014long, henden2020end, coat2021tundra}.

\begin{figure}[h!]
  \centering
  \includegraphics[width=1.0\linewidth, keepaspectratio]{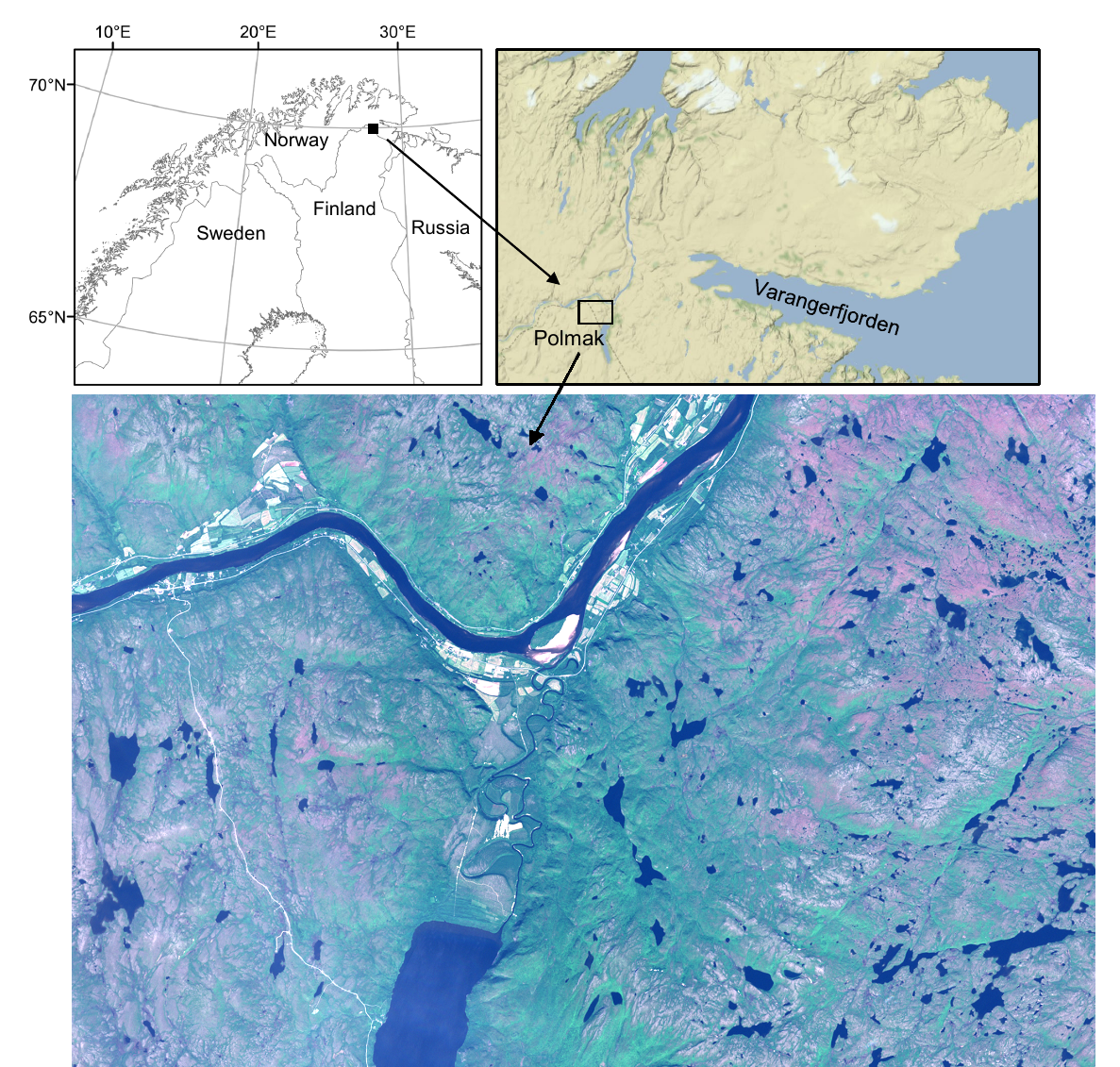}
\caption{A true colour Sentinel-2 image (July 26th 2017) of the Polmak study site, and maps showing its location on the Norwegian-Finnish border. }
\label{fig:aoi}
\end{figure}

The Climate-ecological Observatory for Arctic Tundra (COAT) \parencite{COATscienceplan} is an adaptive, long-term monitoring program, aimed at documenting climate change impacts in Arctic tundra and treeline ecosystems in Arctic Norway.
One of COATs monitoring sites, shown in Figure \ref{fig:aoi}, is located near lake Polmak, partially on Norwegian side and partially on the Finnish side of the border (\ang{28.0}E, \ang{70.0}N).
The chosen study site is subject to different reindeer herding regimes, where the area on the Finnish side of the border is grazed all year round (but mostly during summer), while on the Norwegian side the region is mainly winter grazed \parencite{biuw2014long}.
The site's subarctic birch forest suffered a major outbreak by
both autumnal moth and winter moth between 2006 and 2008, with effects that are still clearly visible in the form of high stem mortality \parencite{biuw2014long}.

Remote sensing imagery is an important tool to observe and understand changes in the \fte{}, both for large scale monitoring and mapping on a local scale.
In this work, we develop a method to find areas with forest mortality after the geometrid moth outbreak, based on satellite images and limited ground reference data.
This is a challenging task for several reasons, where three significant factors stand out and guide our approach to solve the problem.
Firstly, there are few remote sensing images available from our study site.
This is due to the high cloud coverage at high latitudes of subarctic Fennoscandia, which limits the imaging opportunities of optical satellites.
The available cloud-free optical images are relatively few and far between and consecutive images are often from different sensors.
Synthetic aperture radar (SAR) is an active sensor largely uninfluenced by clouds, which can be utilised to monitor defoliation and deforestation \parencite{perbet2019near, bae2021tracking}.
However, planned acquisition of SAR images of the Polmak study site did not start until after the outbreak.
Detecting changes between images from different modalities (e.g.\, SAR and optical), and even between two images from different sensors of the same modality, is very challenging.
If these challenges can be overcome, heterogeneous change detection would enable us to use all available historical data sources for long term monitoring and increase the temporal resolution and responsiveness of the analysis.
The second factor is that changes in canopy state are difficult to detect in medium resolution imagery.
At this scale we do not observe the aggregated landscape level effect as in low resolution satellite images, nor are the individual canopies visible as in high resolution aerial photographs.
For optical images, the loss of "greenness" or normalised difference vegetation index (NDVI) response caused by forest mortality can be offset by the understorey vegetation that becomes increasingly visible as the canopy disappears.
For SAR imagery, the change in scattering mechanisms may help detect forest mortality. However, depending on the forest density, these changes can be very subtle compared to other changes in the scene.
Thus, when using unsupervised change detection methods, the presence of other man-made or natural changes can easily drown out less distinct signs of forest mortality.
Unsupervised methods tend to detect these strong change signatures and attenuate the weaker ones.
Masking of pixels susceptible to such changes by manual inspection, creation of detailed databases of such areas, or pre-classification of images would add another layer of complexity to the task.
The accuracy of the final detection result would also be very dependant on the ability of the masking operation to find all relevant areas, and the spatial resolution would be limited to that of the mask.
Furthermore, it does not prevent other uninteresting changes in vegetation (i.e. not related to canopy state) from appearing in an unsupervised result.
We therefore create a targeted change detection algorithm to learn the change signature for forest mortality based on field data.
The third factor is that learning these signatures with a supervised approach requires labelled data to train the classifier.
While an ecological, field-based, monitoring of forest structure and forest regeneration in the Polmak study site was started in 2011, the scale of the field data is unsuitable to generate training labels for medium-resolution satellite images,
and does not cover the full extent of all ground cover classes contained in the images.
To label data for all classes would be a tedious manual process, and we want to generate just enough training data for our application of detecting forest mortality, since exhaustive classification is unnecessary for our application and could adversely affect the classification accuracy for the class of interest.

We therefore select one-class classification (OCC) to delineate the targeted change in an approach with two main steps:
\begin{enumerate}
    \item Change-aware image-to-image translation that allows direct comparison of heterogeneous pre- and post-event images through differencing.
    \item OCC applied to a stack of difference images (from step 1) and original input images to detect defoliation.
\end{enumerate}

We use the recently developed code-aligned autoencoders (CAE) algorithm \parencite{luigi2020paper3} to do the image-to-image translation.
CAE performs unsupervised change detection. However, since it is based on obtaining the difference images, it can be used to translate images between domains.
Furthermore, since it is designed with change detection in mind, the network learns to preserve the changes in the translations.

\begin{figure}[h!]
  \centering
  \includegraphics[width=1.0\linewidth, keepaspectratio]{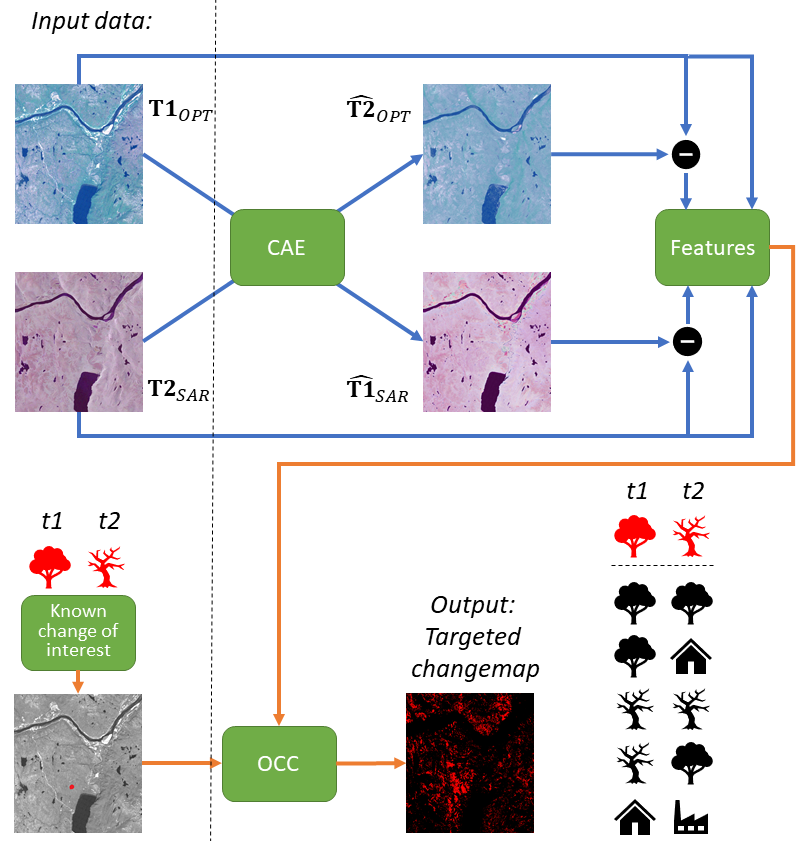}
\caption{Illustration of our approach. The input optical and SAR images at time \textit{t1} and \textit{t2} are translated to the other domain using code-aligned autoencoders (CAE). The originals and the differences with the translated versions are stacked as feature vectors for every image pixel. A limited amount of training data in the form of known areas with forest mortality is then used to train a one-class classifier (OCC) to map the change for a large area in the presence of both unchanged areas and other changes.}
\label{fig:method}
\end{figure}

OCC is a semi-supervised learning approach that utilises the available labels, but does not require a big training set or access to labels for all classes.
By learning the change signature of the phenomenon of interest, we can perform targeted change detection by solving a classification problem with limited and incomplete ground reference data.
For OCC we select a flexible approach that utilises all available ground reference data, i.e.\ also from outside the class of interest. Our approach is summarised in Figure \ref{fig:method}.

The main contributions of this work are:
\begin{itemize}
    \item We propose a method to detect a specific change in heterogeneous remote sensing images based on limited ground reference data and in presence of other changes.
    \item We adapt a deep learning method recently developed for unsupervised change detection to translate the images between domains while preserving the changes.
    \item We adapt a method developed to identify reliable negative samples in OCC for the text domain to work on image data.
    \item We analyse the effect of the number of labelled samples on benchmark datasets and show that relatively little training data is needed to achieve a good classification.
    \item We provide an ablation study for the various components of our method using benchmark datasets.
    \item Our approach allows a post-hoc study of change events, that is study areas that were not originally subject to persistent monitoring, by using any available satellite imagery combination from before and after the event.
    \item Our method allows more responsive detection of changes in areas with high cloud cover.
\end{itemize}

\section{Theory and related work}
\label{sec:theory}

In this section we provide an overview of relevant theory for our application and work related to detection of canopy damage caused by defoliating insects.

\subsection{Remote sensing of insect induced canopy defoliation}
\label{subsec:ecology}

Previous studies of canopy defoliation in the \fte{} have mostly utilised low resolution ($\SI{250}{\metre}$) MODIS data \parencite{jepsen2009monitoring, jepsen2013ecosystem, biuw2014long, olsson2016near, olsson2016development}.
This agrees with the findings in a review of remote sensing of forest degradation by \textcite{gao2020remote}, which shows the prevalence of optical data in general and MODIS in particular.
A literature review by \textcite{senf2017remote} found that studies of disturbances by broadleaved defoliators mainly used low or medium resolution data, with Landsat being the most used sensor.
These works typically use spectral indices and dense time series to detect the defoliation. \textcite{senf2017remote} found that $82\%$ of studies mapping insect disturbance of broadleaved forest used a single spectral index, typically NDVI.
This is consistent with the observation by \textcite{hall2016remote} that image differencing of vegetation indices derived from spectral band ratios were most frequently used.
However, problems with the low resolution of MODIS for mapping forest insect disturbance in fragmented Fennoscandian forest landscape were emphasised by \textcite{olsson2016development}.
Limitations of low spatial resolution sensors for detection of pest damage were also pointed out by \textcite{hall2016remote}, due to the large number of different surface materials that can be contained in a pixel, only some of which are affected by the outbreak.
To rely on a single spectral index makes results susceptible to changes from other sources than forest mortality or dependant on accurate forest masks to avoid this.
The sole use of NDVI also ignores the information contained in other bands of the sensor.

When it comes to the use of SAR data, none of the studies of broadleaf defoliation listed by \textcite{senf2017remote} used SAR, nor did the works on remote sensing for assessing forest pest damage
reviewed by \textcite{hall2016remote}.
A study by \textcite{mitchell2017current} of approaches for monitoring forest degradation did include a number of SAR data applications. However, these were mainly for L- and X-band SAR data.
Most of the summarised papers dealt with the scenario where entire trees (stems and branches) had been removed, for instance by fires or logging, or they used proxy indicators, such as detection of forest roads, to monitor degradation \parencite{mitchell2017current}.
It was found that only a few studies had investigated the use of C-band data \parencite{mitchell2017current}.
The study of C-band SAR remains interesting, particularly because the Sentinel-1 satellites provide free data in that frequency band.

A study of insect induced defoliation using C-band SAR was presented by \textcite{bae2021tracking}, which calculated the correlation between defoliation risk and smoothed time series of backscatter values averaged over five hectare ($\SI{50000}{\metre\squared}$) plots.
In a precursor to this work, we discriminated between live and dead canopy based on an accurate estimation of polarimetric covariance from a single, full-polarimetric C-band image \parencite{agersborg2021guided}.

\subsection{Heterogeneous change detection}
\label{subsec:hetero-cd}

Traditionally, change detection has been based on images from the same sensor and preferably with the same acquisition geometry before and after the change event. This puts some limitations on the use of such methods. Firstly, the response time is at a minimum limited to the revisit time of that particular satellite or constellation. Secondly, the area of interest (AOI) might not be covered frequently by the sensor. Furthermore, the time period which can be studied is also limited to the active time period for that particular sensor.
One way to alleviate these issues is to use heterogeneous change detection on imagery from two different sensors. This comes at the cost of solving a problem that is methodologically more challenging, especially in the unsupervised case, but is still our chosen solution given the practical constraints.

To enable reliable, long-term, persistent monitoring of forest mortality in our AOI, we need to use images from different sensors due to frequent cloud cover.
SAR sensors have significantly higher imaging opportunities because of their near weather-independent nature.
For our AOI, we do not have SAR imagery before the geometrid moth outbreak, and we thus have to use Landsat data for the pre-event image.
Furthermore, we do not have enough data to use time series for smoothing or monitoring gradual changes, as in the studies based on low-resolution MODIS NDVI \parencite{jepsen2009monitoring, jepsen2013ecosystem, biuw2014long, olsson2016near, olsson2016development} or the approach by \parencite{bae2021tracking}. Hence, we must rely on bi-temporal change detection using pairs of images.

Heterogeneous change detection has received growing interest the last years \parencite{touati2019multimodal, sun2021nonlocal}.
Fully heterogeneous change detection should work in a range of settings, from the easiest where the images are acquired with the same sensor, but with different sensor parameters or under different environmental conditions, to the most challenging where images are obtained by sensors that use different physical measurement principles (e.g.\ SAR and optical) \parencite{sun2021nonlocal}.
The advances of deep learning have in recent years opened up several new directions for heterogeneous change detection.
Image-to-image translation in particular offers the interesting prospect of comparing the images directly, once one or both have been translated to the opposite domain.
Many traditional change detection methods involves a step for homogenising the images, such as radiometric calibration, even for images from the same sensor \parencite{coppin2004review}.
Image-to-image translation can be seen as an evolution of this traditional preprocessing step; one (or both) images are "re-imagined" to what it would look like in the other domain.

\subsection{Image-to-image translation}
\label{subsec:im2im}

Code-aligned autoencoders (CAE) is a recently developed method for general purpose change detection designed to work with heterogeneous remote sensing data \parencite{luigi2020paper3}.
The change detection is based on translating images $\preimage$ and $\postimage$ into $\hat{\postimage}$ and $\hat{\preimage}$, respectively, such that the difference images $\preimage-\hat{\preimage}$ and $\postimage-\hat{\postimage}$ can be computed.
The CAE algorithm must ensure that changed areas are not considered when learning the translation across domains to avoid falsely aligning the changed areas.
CAE identifies changed areas in a unsupervised manner and reduces their impact on the learnt image-to-image translation.
The unsupervised nature of the algorithm means that we do not need training data.
While CAE performs change detection directly, it will detect all changes between the two acquisition times, such as those caused by human activity, fluctuating water levels, seasonal vegetation changes, etc., and not just forest mortality.
We therefore utilise it as an image-to-image translation method that does not require labelled training data, is change-aware, and is designed for heterogeneous remote sensing data.

\subsection{One-class classification}
\label{subsec:occ}
Canopy defoliation has a weak \emph{change signature} compared to many other vegetation changes that can be discerned in medium resolution imagery, meaning that the imposed change of the radiometric signal is much smaller than for disturbances such as e.g.\ forest fires or clear cutting. Thus, it will not in general be detected by the CAE.
An unsupervised change detection method will tend to highlight certain strong changes and ignore weaker ones.
If we lower the threshold for detection, the forest mortality could be found, but it would be surrounded and accompanied by many unrelated changes in the final change map.
We therefore use a semi-supervised approach to detect the change phenomenon of interest while ignoring all other changes.

Traditional supervised classification algorithms require that all classes that occur in the dataset are exhaustively labelled \parencite{li2010positive}.
To obtain sufficient training data from all possible classes by manual labelling would be both time consuming and costly \parencite{li2010positive}, and does not necessarily improve the classification accuracy with respect to our class of interest.
While we cannot collect ground reference data for all classes, we still want to utilise the available data to train a change detection algorithm to find changes in a larger region extending beyond the study area.
We instead use techniques from one-class classification (OCC) to learn the change signature from a very limited number of labelled samples of ground reference data containing forest mortality information.

OCC is framed as a binary classification problem, where the class of interest is referred to as the positive class (or target class), with label $\classlabel\!=\!1$.
In this setting, the negative class, $\classlabel\!=\!0$, is either absent from the training data or the instances available "do not form a statistically representative sample" \parencite{khan2014one}.
The negative class is usually a mixture of different classes \parencite{li2020one}, defined as the complement of the positive class.
The typical case for OCC is that the full dataset $\dataset$ is divided into $\posiset$, the set of labelled positive samples, and an unlabelled set $\unlaset$ (also called mixed set) that consists of data from both the positive and the negative class.
OCC seeks to build classifiers that work in such scenarios, which often naturally arise in real world applications \parencite{bekker2020learning}.
In general, the results will not be as good as in true binary classification, where statistically representative samples from both the positive and negative class are available for training.

In our case the reference data is not randomly sampled, but selected systematically from a spatially limited area with a sampling bias that is somewhat related to landscape attributes.
The lack of random sampling limits our choice of OCC algorithm to the so-called \emph{two-step techniques} and discards the other possibilities mentioned in the taxonomy of \textcite{bekker2020learning}.
Two-step techniques only require two quite general assumptions: smoothness and separability \parencite{bekker2020learning}. Smoothness means that samples that are close are more likely to have the same label, while separability means that the two classes of interest can be separated \parencite{bekker2020learning}. The steps are:
\begin{enumerate}
    \item Given labelled positive samples $\posiset$ from a dataset $\dataset$, reliable negative (RN) samples are identified from the remaining set of unlabelled samples $\unlaset$.
    \item A classifier is trained with the provided labelled positive samples $\posiset$ and using the RN samples from Step 1 as the (initial) set of negative training data, $\negaset$.
\end{enumerate}
Any (semi-)supervised classifier can be used in the Step 2, as both positive and RN labels are available.

\section{Methodology}
\label{sec:methods}

\subsection{Feature selection}
\label{subsec:feature-selection}

Feature engineering is an important part of machine learning, as selecting the right features and normalisation may have a big influence on the final classification result.
For our bitemporal change detection problem we originally have the co-registered pre- ($\preimage$) and post-event ($\postimage$) images, which may be from different sensors and even different physical measurement principles.
We want the feature vectors to be as descriptive as possible since we have a very limited amount of training data, and only from the change class of interest.
A common change detection technique for homogeneous images is to subtract one image from the other to obtain the difference image, $\mathbf{D}\in \mathbb{R}^{\imh \times \imw \times \sampledim}$.
The exact steps for finding the changes from the features of $\mathbf{D}$ varies, but often involve a form of thresholding.

We seek to combine original image features and difference vectors for heterogeneous images without creating a specific method for weighting the contributions.
Using CAE for image-to-image translation, we can obtain $\hatpreimage$, which is the post-event image translated to the domain of $\preimage$, and $\hatpostimage$, the pre-event image translated to the domain of $\postimage$.
This allows us to compare the pre- and post-event images and utilise the difference images in our features:
\begin{align}
    \mathbf{D}_\preevent & =  \hatpreimage - \preimage   \label{eq:diff-domain1} \\
    \mathbf{D}_\postevent & = \postimage - \hatpostimage \label{eq:diff-domain2}
\end{align}
If $\preimage$ is an image with $\nchanpre$ channels and $\postimage$ has $\nchanpost$ channels for each pixel position in the $\imsize$ images, Eqs.\ \eqref{eq:diff-domain1} and \eqref{eq:diff-domain2} correspond to the element-wise differences:
\begin{align}
    \mathbf{d}_\preevent & =  \hatprevec - \prevec = [\hat{\preevent}(1) - \preevent(1), \ldots, \hat{\preevent}(\nchanpre) - \preevent(\nchanpre) ]\,,    \label{eq:el-diff-domain1} \\
    \mathbf{d}_\postevent & = \postvec - \hatpostvec = [\postevent(1) - \hat{\postevent}(1), \ldots, \postevent(\nchanpost) - \hat{\postevent}(\nchanpost)]\,, \label{eq:el-diff-domain2}
\end{align}
where $\prevec \in \mathbb{R}^{\nchanpre \times 1}$ and $\postvec \in \mathbb{R}^{\nchanpost \times 1}$ are the multi-channel pixel vectors in the co-registered input images at a given position, $\hatprevec \in \mathbb{R}^{\nchanpre \times 1}$ and $\hatpostvec \in \mathbb{R}^{\nchanpost \times 1}$ are the corresponding pixel vectors in the translated images, and the parentheses are used to index the channels of the image.
For an optical image, the channels will be spectral bands, while for SAR they will typically contain polarimetric information.

By stacking the difference vectors from Eqs.\ \eqref{eq:el-diff-domain1} and \eqref{eq:el-diff-domain2} with the original multichannel input data, we form an array $\samplemat \in \mathbb{R}^{\imh \times \imw \times \sampledim}$ where at each of the $\imh \times \imw$ pixel positions, the feature vector $\sample$ can be written as:
\begin{equation}
\label{eq:feature-vec}
    \sample = [\prevec^T, \mathbf{d}_\preevent^T, \postvec^T, \mathbf{d}_\postevent^T]^T \in \mathbb{R}^{\sampledim \times 1}
\end{equation}
where $(\cdot)^T$ denotes the transpose operation, and the dimension of the feature vector is equal to the sum of the dimension for each component, $\sampledim = \nchanpre+\nchanpre+\nchanpost+\nchanpost = 2 \nchanpre+2\nchanpost$.
The combined feature vector $\sample$ will contain information about both the change, and the state before and after the event.
Since we use labelled training data, we avoid hand-crafted features or dimensionality reduction methods such as PCA, thus allowing the machine learning algorithm to learn which features are important for detecting the change of interest.

\subsection{Building the OCC}
\label{subsec:building-occ}
We select the two-step approach to OCC for our application since it is flexible and makes no assumptions about random sampling of the positively labelled training data.
A further benefit is that enables the use of ground reference data collected from other classes (i.e.\ not forest mortality) by adding it to the RN samples to augment the negative class.
When selecting the methods for each step, we want to avoid those that have many parameters that require tuning.
To further guide our choice, the OCC should work for a range of different number of positive samples.
We also exclude methods that are highly specialised toward the text domain.

\subsubsection{Step 1}
\label{subsubsec:step1}
To obtain RNs in the first step, we use a Gaussian mixture model (GMM) updated once with the expectation maximisation (EM) algorithm \parencite{dempster1977em}. This is inspired by the first step in the Spy algorithm \parencite{liu2002partially} and the well established use of the naïve Bayes (NB) classifier solved with the EM algorithm as the second step \parencite{bekker2020learning}.
Our starting point is the same as for \textcite{liu2002partially}: We want to train a Bayes classifier with the labelled positive data and to use all remaining (unlabelled) samples as negatives, before updating the classifier once using EM.
The classifier uses Bayes' rule, where a data point $\sample$ should be consider a reliable negative if the probability that it belongs to the negative class ($\classlabel=0$) is greater than the probability of belonging to the positive class ($\classlabel=1$):
\begin{equation}
    \prob(\classlabel = 0| \sample) > \prob(\classlabel = 1| \sample).
\end{equation}
These probabilities are then reformulated using Bayes' rule and cancelling the evidence ($\pdf(\sample)$) in the denominators, gives the decision rule:
\begin{equation}
\label{eq:bayes-decision}
    \prob(\classlabel = 0) \pdf(\sample | \classlabel = 0) > \prob(\classlabel = 1) \pdf(\sample | \classlabel = 1)
\end{equation}
where $\prob(\classlabel)$ is the prior probability of the positive or negative class, and $\pdf(\sample | \classlabel)$ is the likelihood.

The approach of NB is to make the so-called \emph{naïve} assumption that all the features of $\sample$ are mutually independent when conditioned on the class $\classlabel$.
Then $\pdf(\sample | \classlabel)$ can be written as the product of the marginal univariate probability density functions (PDFs) of all features in $\sample$.
The use of NB by \textcite{liu2002partially} and other works stems from document classification, where the marginals are modelled as discrete probability mass functions which represent the probability of a given word occurring in a document of class $\classlabel$.
These can be readily estimated by counting word occurrences, while estimating $\pdf(\sample | \classlabel)$ for the set of words in a document is intractable unless the vocabulary is small.
In our case, the naïve assumption of mutual independence of features does not make it easier to calculate Eq.\ \eqref{eq:bayes-decision}, as that requires assumptions about the PDFs $\pdf(\samplefeature_i | \classlabel)$ for each feature $\samplefeature_i \in \sample = [\samplefeature_0, \samplefeature_1, ..., \samplefeature_{\sampledim-1}]$.
Instead, we choose to use a parametric model for the conditional probability density functions for the feature vectors for each class, $\pdf(\sample | \classlabel)$.
Compared to using the naïve approach, this allows us to account for correlation between the features of $\sample$.
Recall that the feature vectors consist of all channels from the original images as well as differences obtained with the translations, as given by Eq.\ \eqref{eq:feature-vec}, so we must assume correlation between features.
Since the goal is to perform classification in order to obtain reliable negative samples which can be used to train a better, final, classifier in the second step, we do not attempt to optimise the selection of $\pdf(\sample | \classlabel)$.
We instead argue that the Gaussian is a reasonable choice of PDF in this setting, since its parameters are readily estimated by the mean and the sample covariance matrix, with the latter capturing the correlation between features.
Thus, we model the marginal density for the positive class as $\pdf(\sample | \classlabel = 1) \sim  \mathcal{N}(\boldsymbol{\mu}_1, \boldsymbol{\Sigma}_1)$, and likewise for the negative class $\pdf(\sample | \classlabel = 0) \sim  \mathcal{N}(\boldsymbol{\mu}_0, \boldsymbol{\Sigma}_0)$.

This is a two-component GMM, and we can use EM to provide an initial classification of the data in order to find RNs.
The initial estimates for the mean and covariance of the first mixture component, $\hat{\boldsymbol{\mu}}_1$ and $\hat{\boldsymbol{\Sigma}}_1$, are based on the labelled positive training set, while the estimates $\hat{\boldsymbol{\mu}}_0$ and $\hat{\boldsymbol{\Sigma}}_0$ are based on the unlabelled set, which contains both positive and negative samples.
The standard sample mean and sample covariance matrix estimators are used.
Using results from EM for GMM (see e.g.\ \textcite{Bishop2006}) we can now find refined estimates for the parameters.
The new estimates of the expectation for mixture component $k$ is then:
\begin{equation}
\label{eq:em-mean}
    \hat{\boldsymbol{\mu}}_k^{\text{new}} = \frac{1}{n_k} \sum_{j=0}^{N-1} \responsibility{jk}  \sample_j  \, \, , \, \, k = 0, 1
\end{equation}
where $\responsibility{jk} \in [0,1]$ is called the responsibility and denotes the posterior probability of sample $\sample_j$ belonging to mixture component $k$, and $n_k = \sum_{j=0}^{N-1} \responsibility{jk}$ is a normalisation factor.
The term indicates how much "responsibility" mixture component $k$ has for explaining sample $j$.
The $\responsibility{jk}$ are calculated as the posterior probabilities of sample $\sample_j$ belonging to mixture component $k$ given the parameters of the mixture components calculated in the previous (initial) iteration of the EM algorithm.
The prior probabilities in Eq.\ \eqref{eq:bayes-decision} are initialised as equal (uninformative) $\pripos = \prineg = 0.5$.
Since we want mixture component $k=1$ to model the positive class, we set $\responsibility{j1} = 1$ and $\responsibility{j0} = 0$ when $\sample_j$ is from the positive training set.
The updated estimate for the covariance matrices are given in a similar manner as Eq.\ \eqref{eq:em-mean} as
\begin{equation}
\label{eq:em-cov}
    \hat{\boldsymbol{\Sigma}}_k^{\text{new}} = \frac{1}{n_k} \sum_{j=0}^{N-1} \responsibility{jk}  (\sample_j - \hat{\boldsymbol{\mu}}_k) (\sample_j - \hat{\boldsymbol{\mu}}_k)^T   \, \, , \, \, k = 0, 1 .
\end{equation}

We note that the EM estimates are the maximum likelihood estimators (MLEs) for the expectation and covariance matrix weighted by $\responsibility{jk}$.
These can be easily found by functions that can calculate weighted versions of the mean and sample covariance matrix, e.g.\ the \textit{average} and \textit{cov} functions from the Python \texttt{numpy} library.
The reliable negatives are selected as the samples $\mathbf{x}$ where the probability of belonging to mixture component used to model the negative class is greater than that of the positive class according to Eq.\ \eqref{eq:bayes-decision}.
The prior probabilities are calculated as the proportion of samples assigned to each class and the estimates in Eqs.\ \eqref{eq:em-mean} and \eqref{eq:em-cov} for the positive and negative class are then used to evaluate the Gaussian PDFs.
We used the \textit{multivariate\_normal} function from the \texttt{scipy.stats} package.
Intuitively, the mixture component for the negative class will be "wide" and have the largest value for most of the support, while the mixture component of the positive class will be "compact" and only have higher values close to where the positive training samples are located.

\subsubsection{Step 2}
\label{subsubsec:step2}

For the second step, any (semi-)supervised method could be used, as we now have training data for both classes with the labelled positives and RNs \parencite{bekker2020learning}.
We choose to base our second step on multi-layer perceptron (MLP) feed-forward neural networks.
Neural networks are a good out-of-the-box method for a problem such as this, due to their high flexibility as function approximators.
Unfortunately, there are no general guidelines for selecting the number of hidden network layers, or the amount of neurons in each.
Our initial data exploration revealed that there were some variations in the classification results depending on how the network architecture was selected.
We therefore opted to combine five different MLPs in an ensemble model, and use the majority vote to determine the class of each pixel.
The architectures of the ensemble consisted of one MLP with a single hidden layer of $1000$ neurons, two MLPs with two hidden layers, one with $100$ and the other with $200$ neurons in both layers, and two MLPs with three hidden layers, again with uniform layer size of either $100$ or $200$ neurons in all layers.
Except for the architecture,  all MLPs used the default parameters for the \textit{MLPClassifier} function from the Python  Scikit-learn \texttt{sklearn.neural\_network} package \parencite{scikit-learn}.
The default parameter selection uses the rectified linear unit (ReLU) activation function and the Adam optimiser \parencite{kingma2014adam}.
The ensemble setup and MLP parameters are kept constant for all experiments.
After training, the ensemble of MLPs is used on the full dataset to find the targeted change.

\section{Results}
\label{sec:results}

\subsection{Illustrating targeted change detection}
\label{subsec:targeted-cd}
To illustrate targeted change detection, we test our method on a dataset used in the change detection literature.
The Texas dataset consists of two $1534 \times 808$ pixel multispectral optical images of Bastrop County, Texas, where a destructive wildland-urban fire struck 4 September 2011 \parencite{volpi2015spectral}.
The pre-event image is from Landsat 5 Thematic Mapper (TM) with $7$ spectral channels and the post event image is from Earth Observing-1 Advanced Land Imager (EO-1 ALI), with $10$ spectral channels.
The ground truth was provided by \textcite{volpi2015spectral}.

\begin{figure*}[h!]
  \centering
        \begin{subfigure}[b]{0.24\textwidth}
            \centering
            \includegraphics[width=\textwidth]{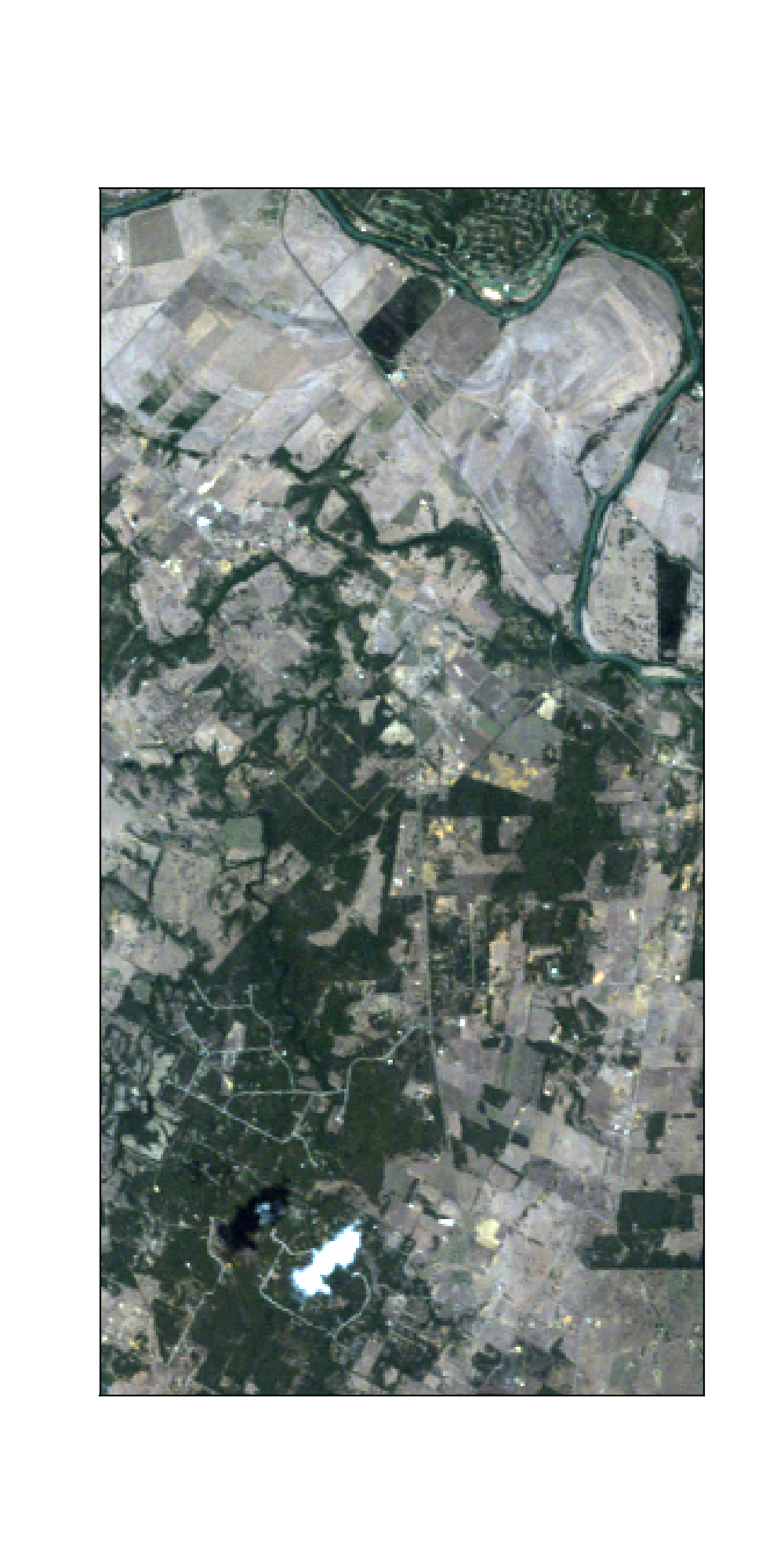}
            \caption[]%
            {{\small Pre-event}}
            \label{subfig:texas-pre-rgb}
        \end{subfigure}
        \begin{subfigure}[b]{0.24\textwidth}
            \centering
            \includegraphics[width=\textwidth]{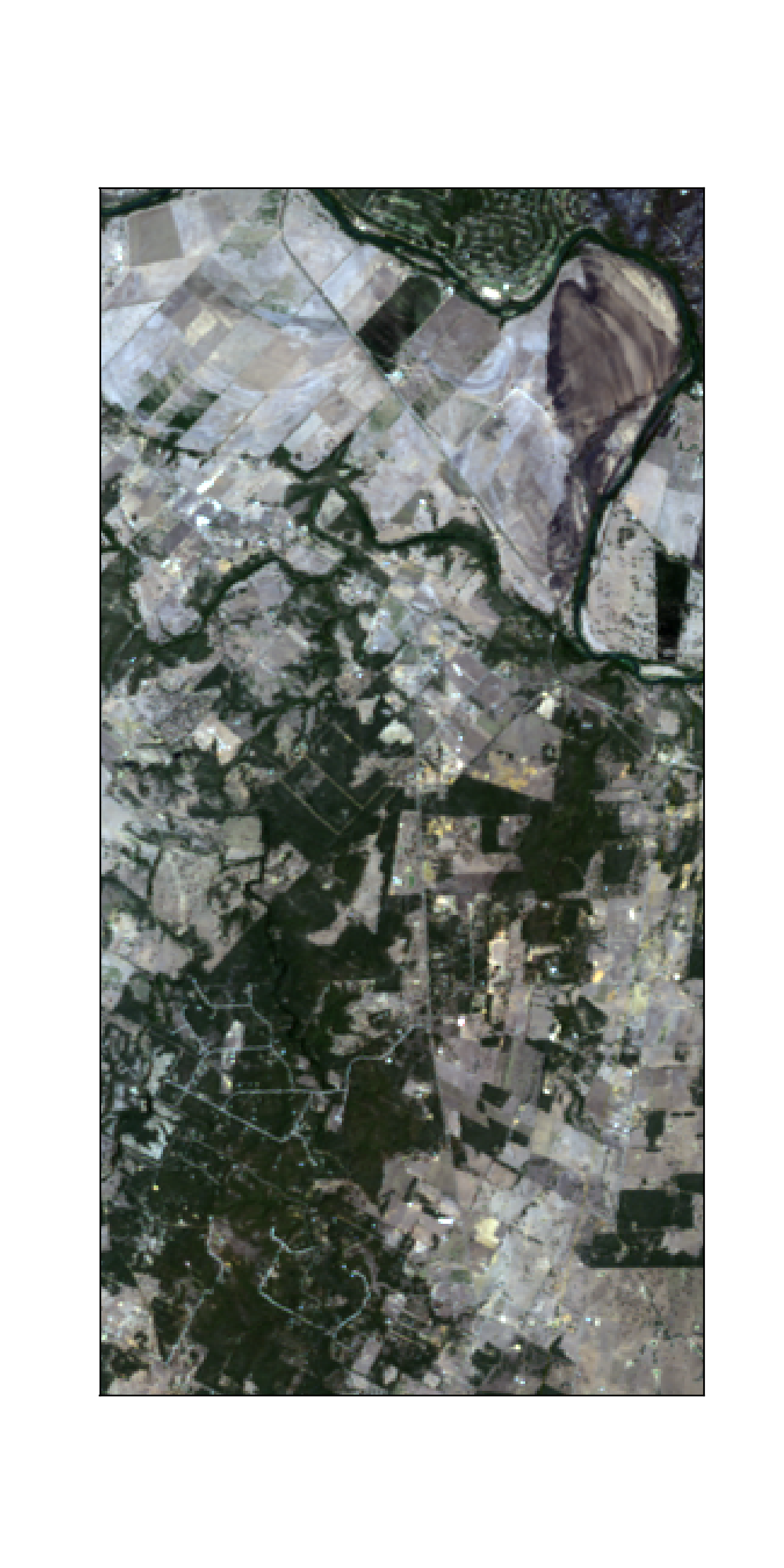}
            \caption[]%
            {{\small Post-event}}
            \label{subfig:texas-post-rgb}
        \end{subfigure}
        \begin{subfigure}[b]{0.24\textwidth}
            \centering
            \includegraphics[width=\textwidth]{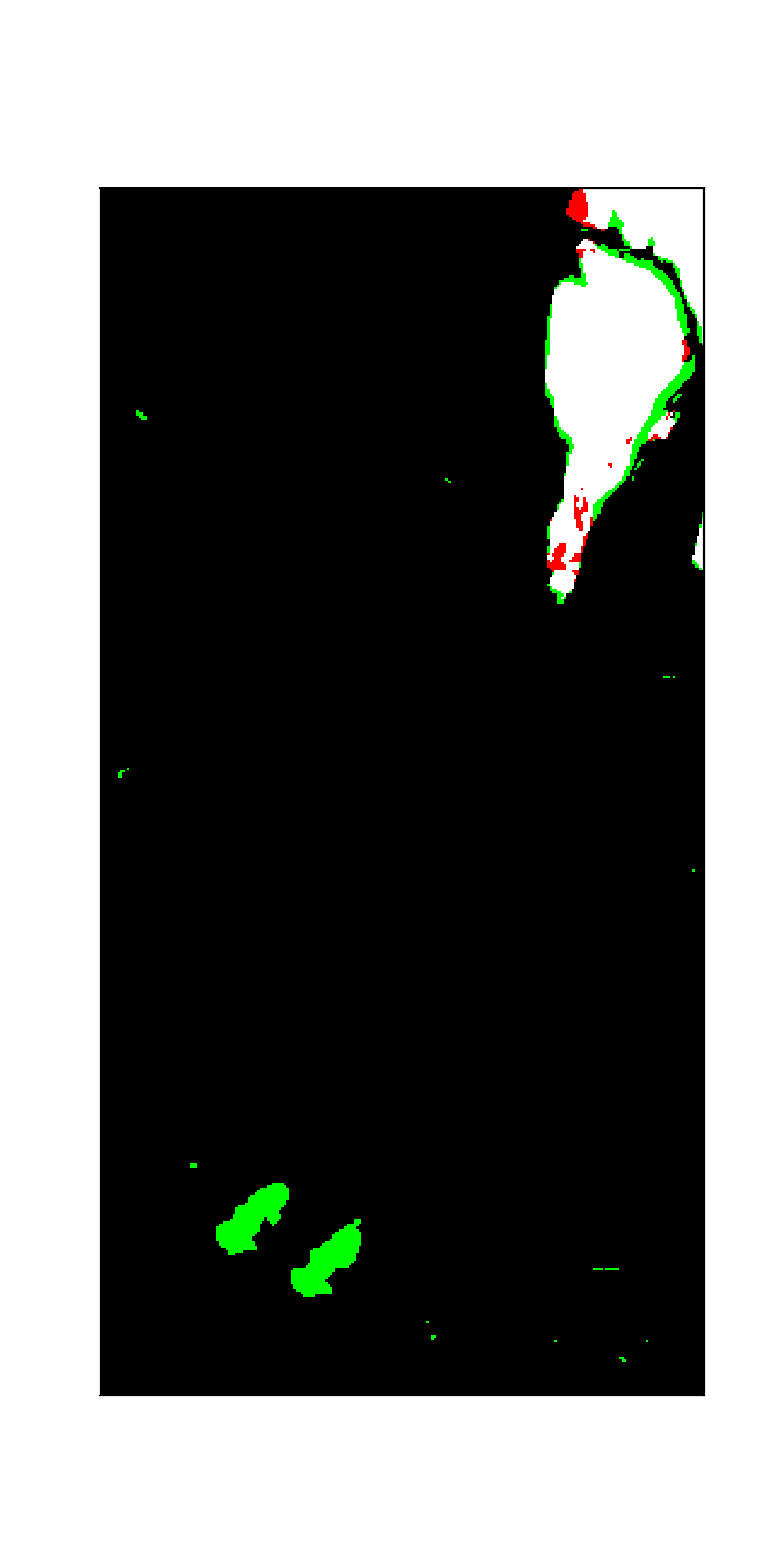}
            \caption[]%
            {{\small CAE}}
            \label{subfig:texas-cae-confmap}
        \end{subfigure}
        \begin{subfigure}[b]{0.24\textwidth}
            \centering
            \includegraphics[width=\textwidth]{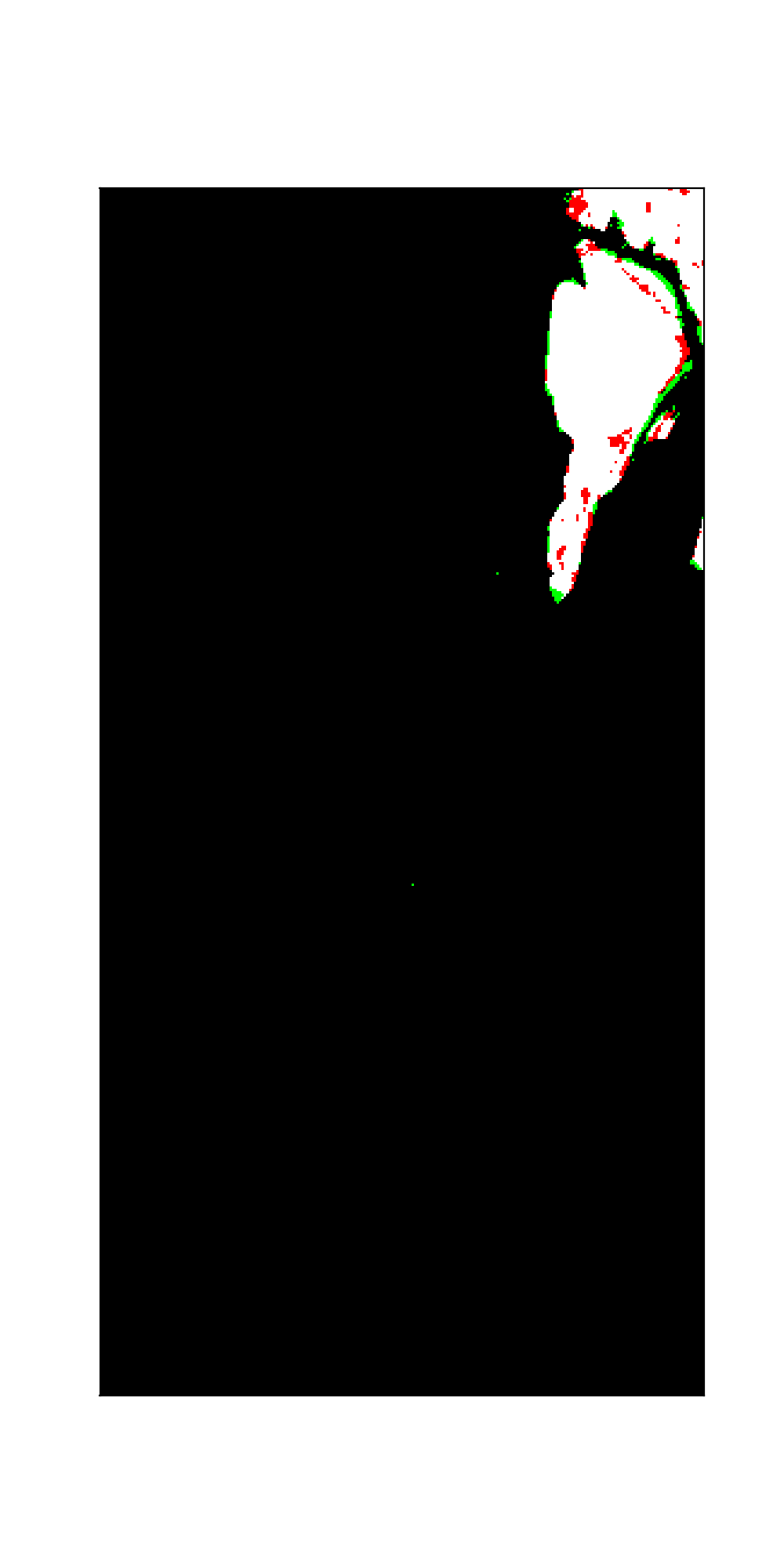}
            \caption[]%
            {{\small OCC}}
            \label{subfig:texas-occ-confmap}
        \end{subfigure}
        \caption[Texas clouds.]
        {\small The spectral bands corresponding to red, green, and blue (RGB) for the two images are shown in the two leftmost subplots. The two rightmost plots show the confusion maps for the unsupervised CAE result and our OCC method, where white pixels represent true positive (TP) classifications, black true negative (TN), green false positive (FP), and red false negative (FN).
        The small clouds and their shadows in the Landsat 5 image are detected as changes compared to the EO-1 image in the unsupervised CAE change detection result, and appear pairwise as green areas (FP) in the confusion map.}
\label{fig:texas-comparison}
\end{figure*}

We apply our method with $1000$ randomly drawn positive samples as the training data.
This corresponds to $0.76\%$ of the positive ground truth data ($0.08\%$ of the total image pixels).
In Figure \ref{fig:texas-comparison} we zoom in on an area containing both the targeted change (where the fire has occurred) and other changes (clouds), and show the original image data and the change detection result.
In addition to the result from our approach, we show that of the unsupervised CAE change detection from \textcite{luigi2020paper3}.
This illustrates how our semi-supervised OCC-based approach ignores changes not present in the labelled training set.
The result in Figure \ref{subfig:texas-cae-confmap} is reasonable, since the clouds (and their shadows) represent an actual change between the two acquisition times.
However, it is the effects of the forest fire which is interesting for us to map, and thus the cloud-related changes are marked as false positives in the ground truth data.

\subsection{Ablation study}
\label{subsec:ablation}
We perform an ablation study to check the contributions of the various components of our approach.
In this study we remove or replace a component and assess the result when using the ablated procedure.
The objective is to check if all components contribute positively to the change detection, or if a simpler method performs equally well or better.
This is an important exercise in the field of machine learning due to the complexity of the methods involved.
Since we wish to objectively measure the effects of the ablation, we must use benchmark datasets where the ground truth is available and we can evaluate the performance of our method numerically.
As a part of this study we also investigate the effect of the number of labelled positive training samples, and how many are needed for a stable result.
We use three benchmark change detection datasets with heterogeneous pre- and post-event images for which ground truth is available.
Though these datasets are not directly related to the task of mapping forest mortality, this study allows us assert that our approach is valid for problems beyond our AOI.
We present how we performed the study before discussing the result for one of the benchmark datasets in detail, and then briefly summarise the two others.

The labelled positive training data is created by randomly drawing a number of positive samples from the ground truth,
and we vary the number used between $25$ and $3000$.
For each positive sample set we use it as training data for the two-step classifier
and calculate the F1 score of the classification result.
The F1 score is defined as the harmonic mean of precision and recall and is a popular metric for evaluating the performance of binary classifiers.
Intuitively, the F1 score puts equal weight on the positive predictions being precise (few false detections) and on finding all positive samples.
Due to its emphasis on both these aspects, the F1 score gives a better characterisation of classifier performance than the traditional overall accuracy measurement, particularly when the classes are unbalanced.
Expressed in terms of true (T) and false (F) classification of positives (P) and negatives (N), the F1 score is given as:
\begin{equation}
    \label{eq:f1}
    \text{F1} = \frac{\text{TP}}{\text{TP} + 0.5 (\text{FP} + \text{FN})}
\end{equation}
where we see that $\text{F1} \in [0, 1]$.
For each training set size, $\npos$, we repeat the experiment $10$ times with a different random sample and calculate the mean F1 score.
We keep all samples from the preceding set as the number of labelled positives is increased to keep the result consistent.

We compare our proposed approach to five different alternatives, three of which are straightforward ablations of our method; one where we drop the second step and two with a reduced feature vector.
Compared to our proposed feature vector $\sample$ in Eq.\ \eqref{eq:feature-vec}, the two feature-related ablations are: not including the differences $\mathbf{d}_\preevent$ and $\mathbf{d}_\postevent$, and not including the original image pixel vectors $\prevec$ and $\postvec$.
The alternative difference vectors are then $\sample_{\text{w/o differences}} = [\prevec^T, \postvec^T]^T$ and  $\sample_{\text{w/o originals}} = [\mathbf{d}_\preevent^T, \mathbf{d}_\postevent^T]^T$.
We also include the F1 score for the GMM with one EM update used to find the reliable negatives (RNs) in the first step.
The difference between this result and our proposed method is the contribution of the second step to the final OCC result.

We also consider an alternative method in the second step, which uses the same RNs as our proposed approach.
The method is called iterative support vector machine (SVM) \parencite{bekker2020learning} and is a common choice for step two.
It successively trains SVM classifiers based on the labelled positive and RN samples available, and then classifies the remaining unlabelled dataset.
The samples which are classified as negative are added to the RN set and used to train a new SVM in the next iteration.
After some convergence criterion is met, the final SVM trained is used to classify the entire dataset.
Due to the large number of RNs found in the first step, we cannot use kernel SVM and must use the linear variant.
The SVM penalty parameter (see e.g.\ \textcite{Bishop2006}) is set by an automated search procedure, as recommended by \textcite{hsu2003practical}, to select the best value based on the false negative rate (FNR).
The iterative SVM  training stops when the FNR exceeds $5\%$, as was first proposed by \textcite{li2003learning}.
Both the MLP ensemble and the iterative SVM use the same RNs found in the first step by the GMM with one EM update.

Finally, we also include results for the popular one-class SVM (OCSVM) method \parencite{khan2014one}.
While it is not an ablation of our approach, OCSVM is an alternative to the first-step method we use, and it is also a frequently used OCC method on its own.
Furthermore, in contrast to the iterative SVM method, the training is only based on the relatively few labelled positive samples, which enables the use of kernel methods.
We use the Scikit-learn implementation of OCSVM, \textit{OneClassSVM} from \texttt{sklearn.svm}, with default parameter values, which uses a RBF kernel with kernel size determined by the number of features and sample variance.

\begin{figure}[h!]
  \centering
  \includegraphics[width=1.0\linewidth, keepaspectratio]{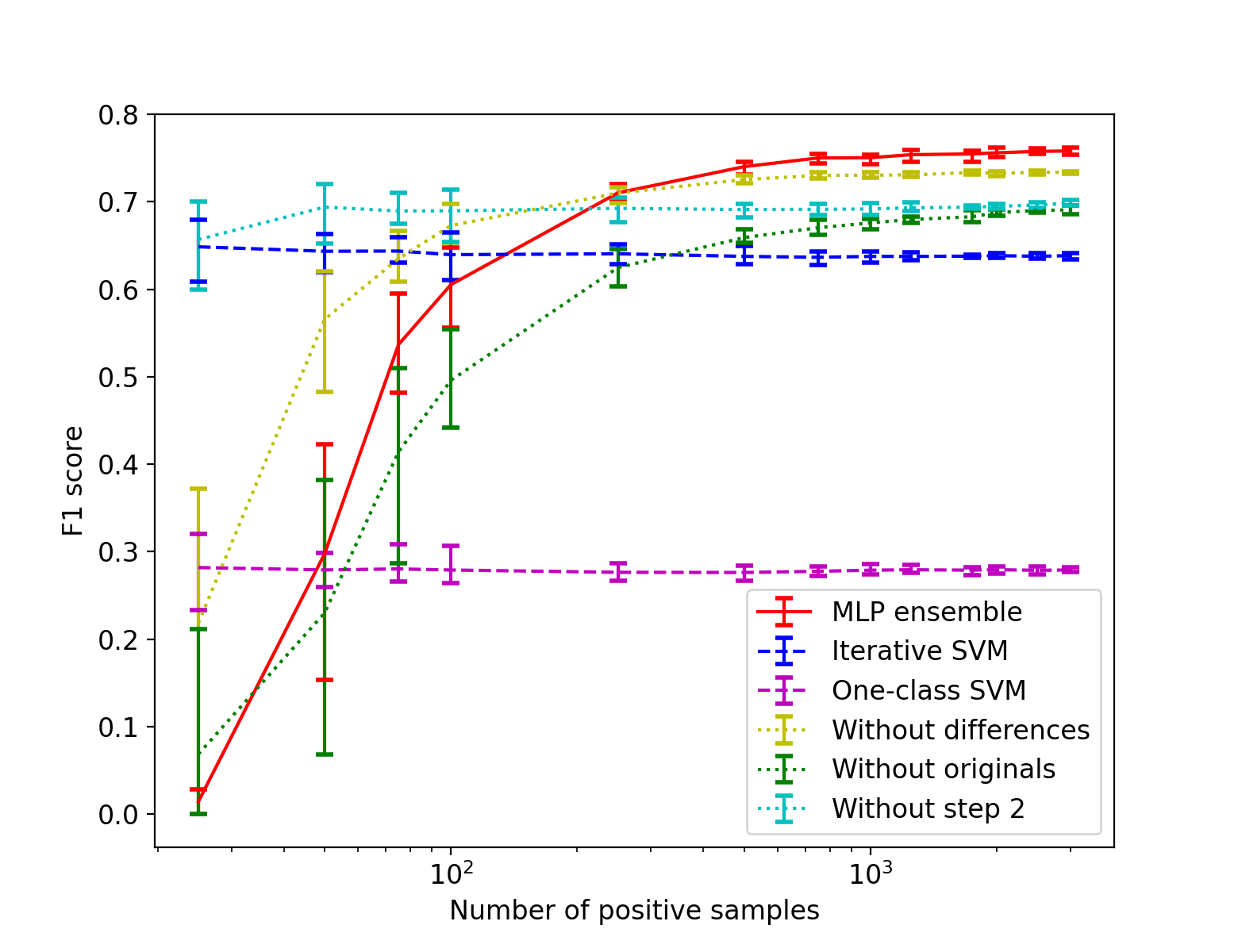}
\caption{F1 scores for all the ablations and methods tested with the France dataset and for different positive training set sizes (logarithmic). }
\label{fig:france-f1}
\end{figure}

We plot the F1 scores as a function of the number of positive samples, $\npos$, for a heterogeneous change detection dataset consisting of two optical RGB images acquired by Pleiades and WorldView 2 in May 2012 and July 2013, respectively. The ground truth is provided by \textcite{touati2019multimodal}. The dataset shows construction work occurring in Toulouse, France, and the list of changes includes earthwork, concrete laying, building construction, and more.
The areas that have been changed were a mixture of different landcover types in the pre-event image, including bare soil, urban, and vegetated.
Figure \ref{fig:france-f1} shows the average F1 score of the $10$ runs plotted as a function of increasing number of labelled positive training samples on a logarithmic x-axis.
The error bars represent the 10th and 90th percentile F1 score.
We see that the iterative SVM second step performs well when the number of labelled positive samples is small. However, the algorithm actually decreases the F1 score from the first step.
The OCSVM has relatively consistent performance as a function of $\npos$, but a considerably lower F1 score than the GMM with one EM update used in the first step.
The OCSVM was originally designed for anomaly detection, a setting where there often is no unlabelled (or negative) data available.
The poor performance in this setting compared to our GMM EM method illustrates the importance of utilising the unlabelled data.
The ablation of difference features performs better than the ablation of original features, and also better than with the full feature vector when $\npos$ is small.
Our proposed approach has the best F1 score when $\npos \geq 250$.

In Table \ref{tab:ablation} we list the F1 scores for the different ablations when $\npos = 1000$. In addition to the result for the France dataset, which corresponds to the values for one x-value in Figure \ref{fig:france-f1}, we include the result for two additional heterogeneous change detection datasets where ground truth information is available.
One is the Texas dataset used for the example in Figure \ref{fig:texas-comparison}.
The other dataset concerns mapping the extent of a lake overflow in Sardinia, Italy. Both images are obtained by Landsat-5, with the pre-event image obtained in September 1995 consisting of a single channel: the near infrared (NIR) spectral band. The post-event image contains the RGB channels and is from July 1996. The ground truth provided by \textcite{touati2019multimodal}.
\begin{table}[h]
    \begin{center}
        \begin{tabularx}{\linewidth}{ c c c c c c c c }
            \hline
            Dataset & W/o originals & W/o differences & W/o step 2 & OCSVM & ISVM & MLP ensemble  \\
            \hline
            France        &  0.676 & 0.730  & 0.692 & 0.279 & 0.638 & 0.750  \\
            Texas         &  0.804 & 0.940  & 0.869 & 0.635 & 0.965 & 0.951  \\
            Italy         &  0.729 & 0.664  & 0.581 & 0.503 & 0.607 & 0.782  \\
            \hline
        \end{tabularx}
    \end{center}
    \caption{F1 score for 1000 postitive samples.}
    \label{tab:ablation}
\end{table}

Table \ref{tab:ablation} shows that both the originals and the differences are important to include in the feature vectors.
The original image has the biggest contribution for the France and Texas datasets, while the difference vectors are most important for Italy.
For all datasets the best result is achieved with the full feature vectors.
The GMM with one EM update performs much better than OCSVM.
For the second step our MLP ensemble improves the F1 score significantly compared to the first step for all datasets.
The ISVM results are more varied.
It decreases the F1 score for the France dataset, as seen in Figure \ref{fig:france-f1}, but achieves a slightly higher F1 score than the MLP ensemble for the Texas dataset.

We conclude that all components of our method contribute to the F1 score.
Relatively few positive samples are needed for a stable classification result, which bodes well for our forest mortality classification.
We also note that our method performs well for these datasets despite that the changed areas in the pre-event images of the benchmark datasets are more varied than for forest mortality mapping. Several different landcover types are affected by the construction, fire, and flooding, whereas for our application the changed areas are always live forest in the pre-event image.

\subsection{Creating the forest mortality map}
\label{subsec:evaluation}

There are few cloud-free optical images available from our AOI, which limits the selection of imagery which could be used to map the forest mortality that has occurred.
We found one Landsat-5 (LS-5) Thematic Mapper (TM) image from 3 July 2005 reasonably close to the start of the geometrid moth outbreak (2006) which we use as the pre-event image.

For the post-event image we use a fine-resolution quad-polarisation RADARSAT-2 (RS-2) scene from 25 July 2017.
We first performed radiometric calibration and terrain correction with the GETASSE30 digital elevation model (DEM) in the Sentinel Application Platform (SNAP), keeping the data in single-look complex (SLC) format with $\SI{10.0}{\metre} \times \SI{10.0}{\metre}$ spatial resolution.
Then the polarimetric covariance matrices were estimated using the guided nonlocal means method \parencite{agersborg2021guided}.
This method preserves SLC resolution and was shown to give estimates of the polarimetric features that better separate between live and dead canopy than alternative methods \parencite{agersborg2021guided}.
The features relevant for canopy state classification were extracted from the covariance matrix $\mathbf{C}$.
These are the intensities in the HH, HV, and VV channels, $C_{11}$, $C_{22}$, $C_{33}$ respectively, and the cross-correlation between the complex scattering coefficients for HH and VV, $C_{13}=|C_{13}|e^{j\angle C_{13}}$ \parencite{agersborg2021guided}.

The Landsat-5 TM bands were upsampled from the original $\SI{30.0}{\metre}$ ($\SI{120.0}{\metre}$ for Band 6) spatial resolution to give a pixel size of $\SI{10.0}{\metre} \times \SI{10.0}{\metre}$ using bilinear interpolation resampling during the coregistration process with the SAR data.
In this process, both the LS-5 and RS-2 images were geocoded to the UTM 35N projection and mapped on a common grid using QGIS before cropping and extraction of overlapping images. This resulted in $1399 \times 1278$ pixel images which were the input for the analysis.

The training data was created by experts carefully comparing high-resolution aerial photography from before (2005) and after the outbreak (2015), drawing polygons covering areas with forest mortality.
We choose this approach for three reasons.
Firstly, it was easier than selecting a greater number of smaller areas from all over the scene, especially considering that the aerial photographs only covered parts of it.
The areas need to be relatively large as the original $\SI{30.0}{\metre}$ resolution of the LS-5 data sets a lower limit for the polygon size.
Secondly, by extracting from within larger homogeneous areas we minimise the effect of any misalignment between the ground reference data and the satellite imagery, which could cause the training data to contain pixels from the negative class.
Thirdly, manual creation of training data is tedious work, and we want to generate just enough training data for our OCC method to map the forest mortality for the entire scene.
15 polygons of roughly rectangular shape with various sizes were created.
Four of these intersects with transects studied in field work in 2017 \parencite{agersborg2021guided}.
In total the 15 polygons contained $1536$ pixels, which given the excerpt size of $1399 \times 1278$ constitutes $0.086\%$ of all pixels.

\begin{figure*}[p]
        \centering
        \begin{subfigure}[b]{0.475\textwidth}
            \centering
            \includegraphics[width=\textwidth]{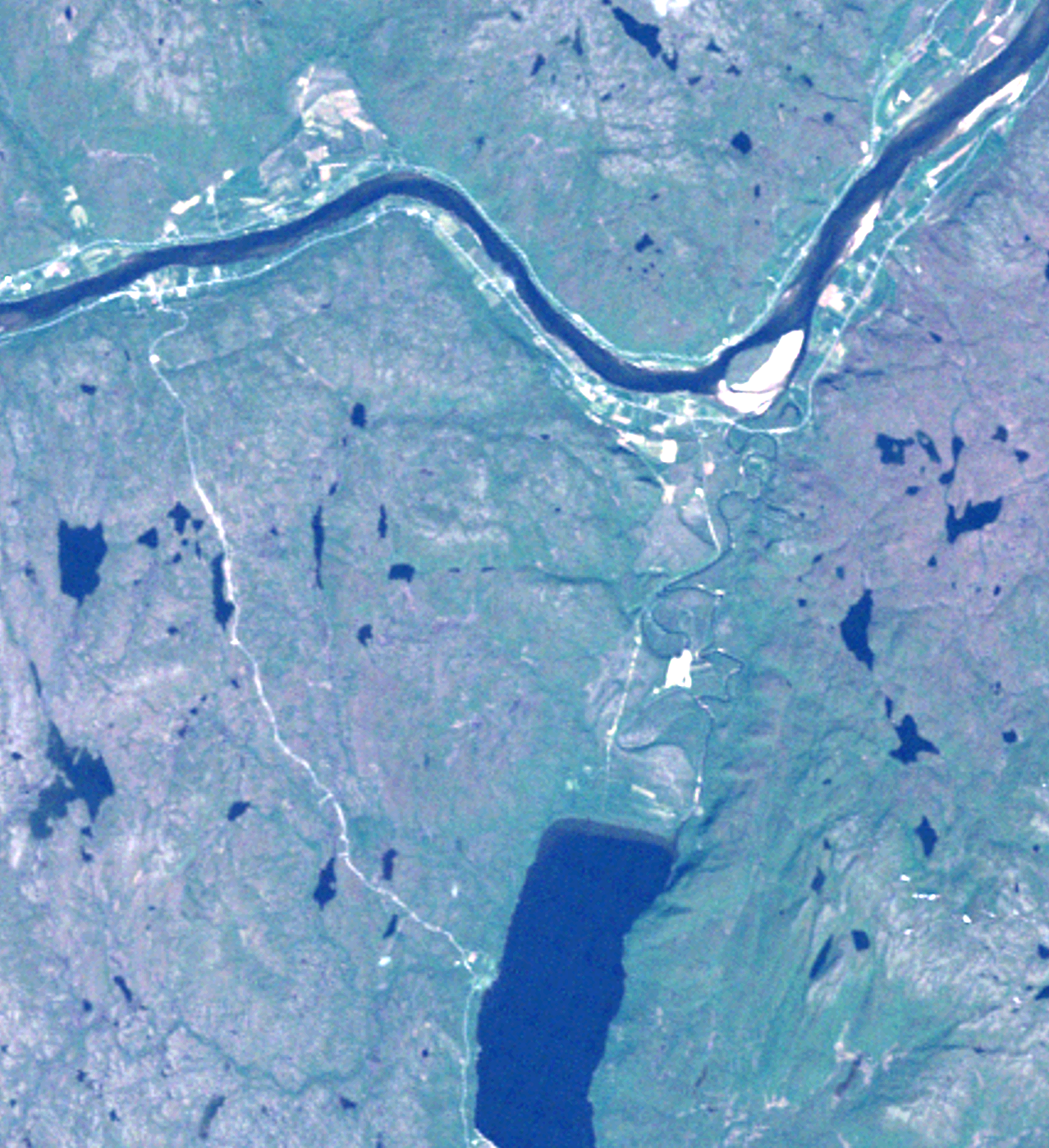}
            \caption[LS-5]%
            {{\small LS-5}}
            \label{subfig:cm2-org-LS2005}
        \end{subfigure}
        \hfill
        \begin{subfigure}[b]{0.475\textwidth}
            \centering
            \includegraphics[width=\textwidth]{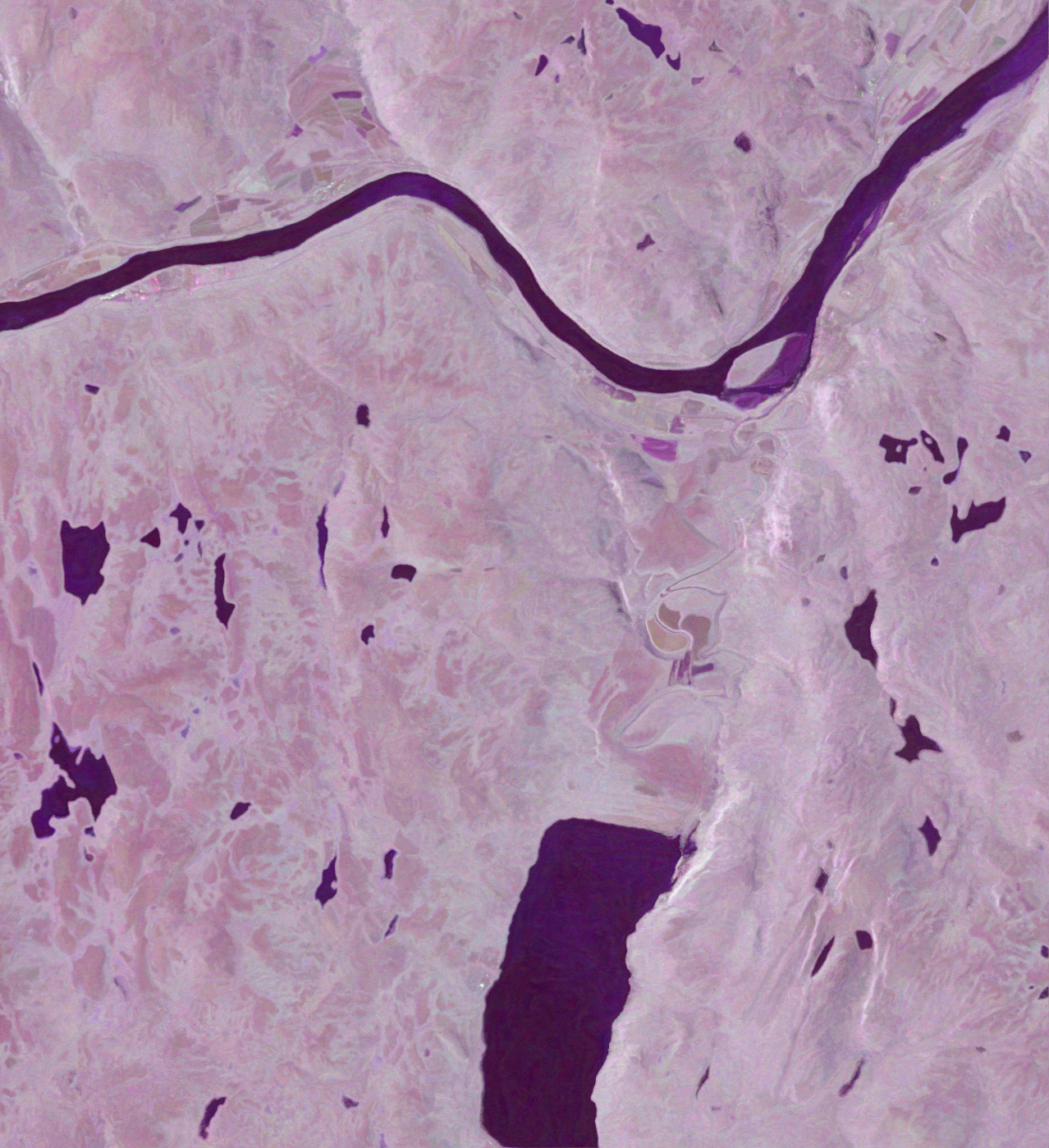}
            \caption[]%
            {{\small RS-2}}
            \label{subfig:cm2-org-PGNLM_A}
        \end{subfigure}
        \vskip\baselineskip
        \begin{subfigure}[b]{0.475\textwidth}
            \centering
            \includegraphics[width=\textwidth]{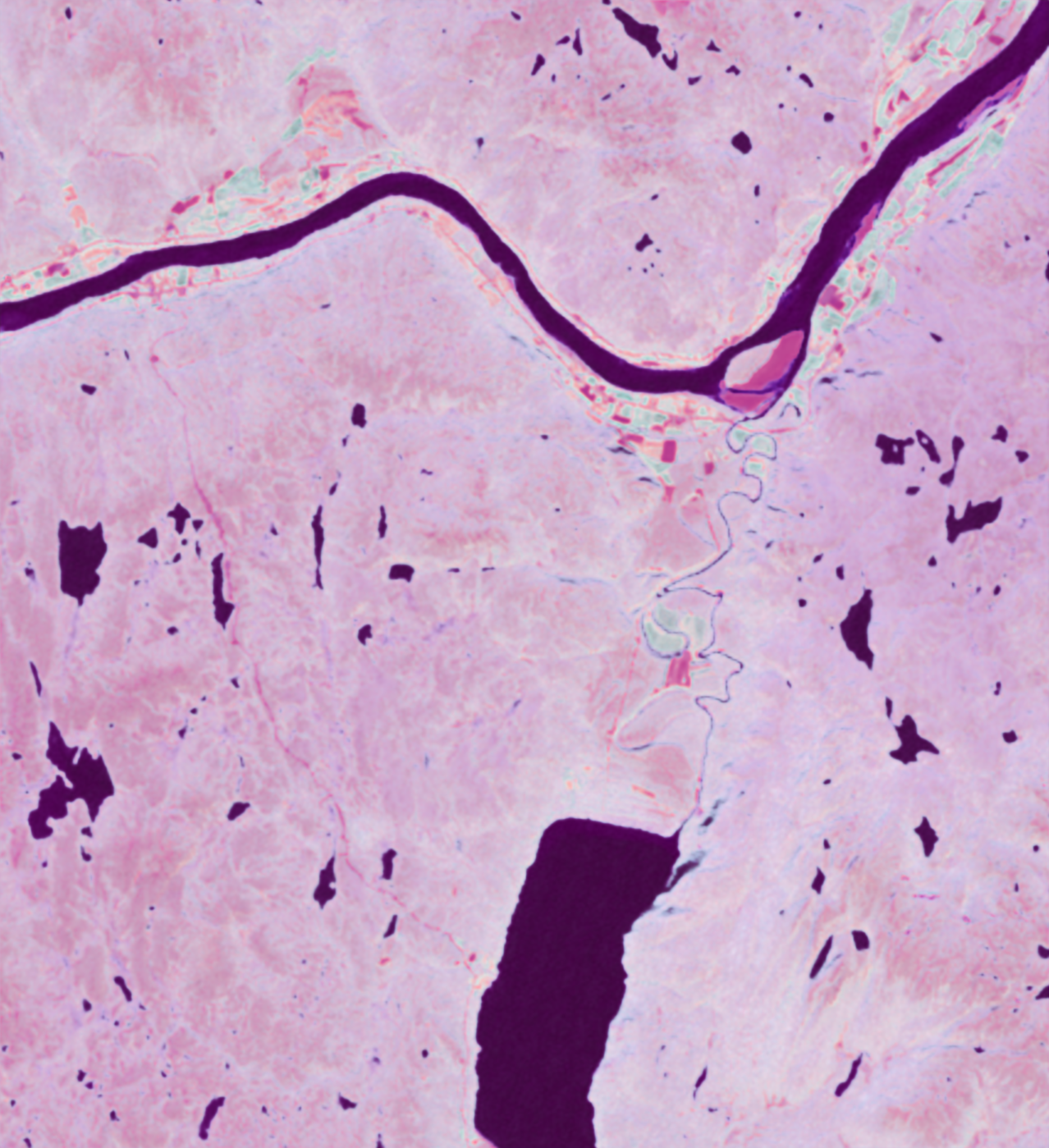}
            \caption[]%
            {{\small Translated LS-5}}
            \label{subfig:cm2-trans-LS5-SAR}
        \end{subfigure}
        \hfill
        \begin{subfigure}[b]{0.475\textwidth}
            \centering
            \includegraphics[width=\textwidth]{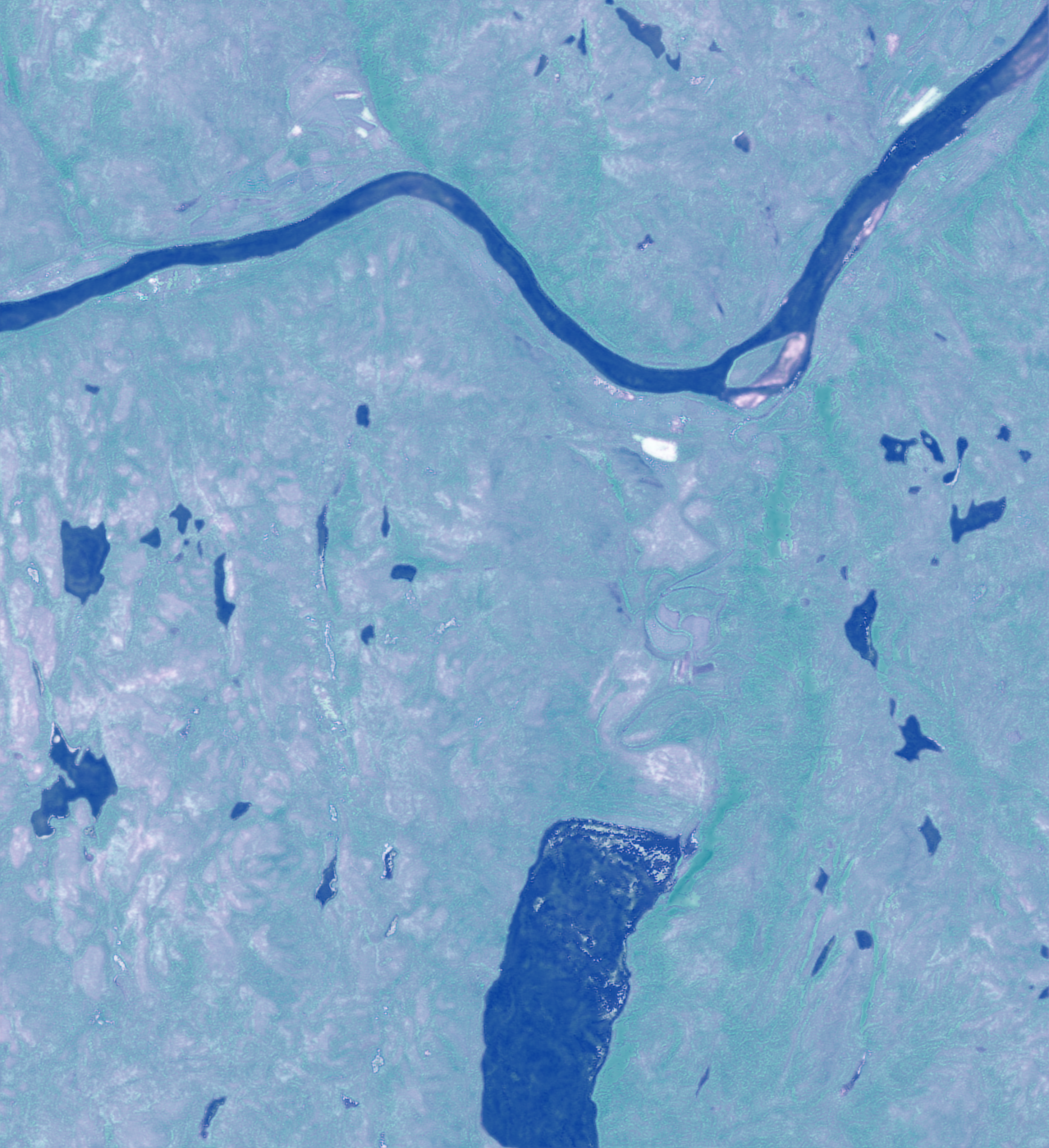}
            \caption[]%
            {{\small Translated RS-2}}
            \label{subfig:cm2-trans-A-OPT}
        \end{subfigure}
        \caption[ Data and translation]
        {\small Dataset pair and translation. Top left: Landsat-5 RGB image. Top right: RADARSAT-2 HH, HV, and VV intensities. Bottom left: Translation of LS-5 image with same channels as \ref{subfig:cm2-org-PGNLM_A}. Bottom right: Translation of RS-2 image with same channels as \ref{subfig:cm2-org-LS2005}. }
        \label{fig:CG2-a-five}
\end{figure*}

Figure \ref{fig:CG2-a-five} shows the satellite images and the corresponding CAE translations. The red, green, and blue bands (Band 3, 2, and 1) of the Landsat-5 TM image are shown in the corresponding RGB channels.
For the RADARSAT-2 scene we use the intensities for the HH, HV, and VV polarisations as the red, green, and blue channels respectively.
The translations are shown below the corresponding original using the RGB channels as the other domain.
Noteworthy features in the image include lake Polmak, in the centre of the lower half of the image, and the Tana river, which runs approximately from east to west in the upper half of the image.
There is also a dirt road in the leftmost part of the image going from the Tana river and south towards the western bank of lake Polmak, which is clearly visible in the optical imagery, but very hard to discern in the SAR data.

The CAE translations in Figure \ref{fig:CG2-a-five} are reasonably similar to the other image in the translated domain, but can easily be discerned from the original data.
This is expected as it is not possible to exactly recreate the spectral information found in optical imagery from SAR data, and likewise, the polarimetric information about scattering characteristics cannot be replicated from the Landsat TM reflectance measurements.
The translation from the optical to the SAR domain seems to match the original RS-2 image quite well, and appears visually to be the better of the two translations, whereas the translation in \ref{subfig:cm2-trans-A-OPT} appears somewhat blurry with muted colours.
There are some obvious translation artefacts, for instance the bright pixels in lake Polmak in the translated RS-2 result.

\subsubsection{Unsupervised change detection with CAE}
\begin{figure}[h!]
            \includegraphics[width=\linewidth]{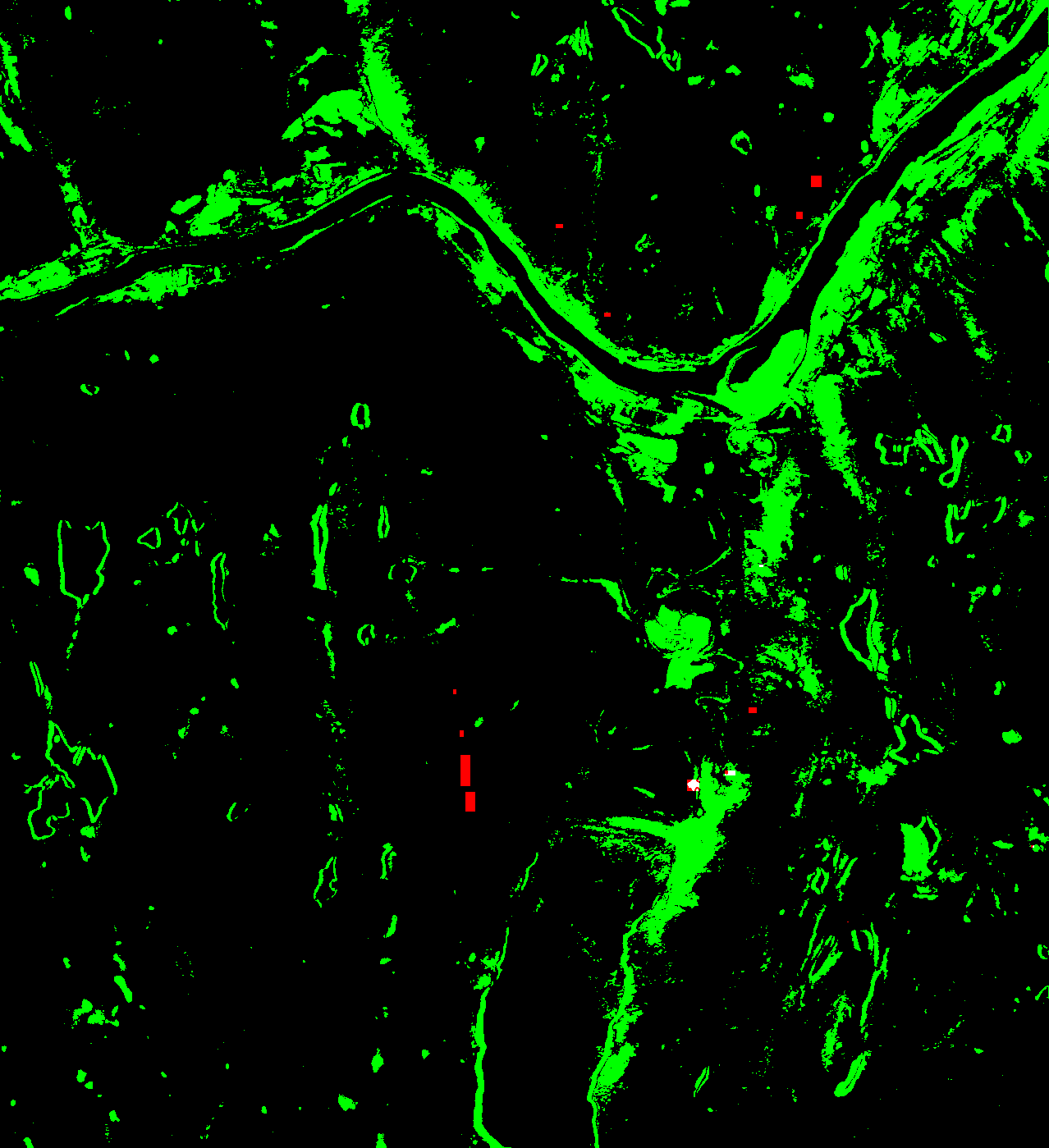}
        \caption[ CAE confusion map.]
        {\small CAE confusion map, with predicted changes in green, predicted unchanged in black, correctly classified forest mortality from our limited ground reference dataset marked white, and red showing missed detections from the same dataset.}
        \label{fig:cae-confmaps}
        \label{subfig:cae-confmap-PGNLM_A}
\end{figure}

To illustrate the changes detected by a non-targeted approach based on the translations in Figure \ref{fig:CG2-a-five}, we show the CAE result.
Change detection with CAE is based on thresholding the magnitude of the difference vectors between the translations and the originals \parencite{luigi2020paper3}.
Figure \ref{subfig:cae-confmap-PGNLM_A} shows the confusion map for the CAE change detection based on the difference vectors obtained from the images shown in Figure \ref{fig:CG2-a-five}.

Figure \ref{fig:cae-confmaps} also shows the 15 ground reference polygons with forest mortality created as training data for the OCC.
Note that some of the polygons are too small to be discerned in Figure \ref{fig:cae-confmaps}.
Parts of the forest mortality polygons where CAE predicts no change are shown in red, while correct change predictions for these areas are white in Figure \ref{fig:cae-confmaps}.
Only $14\%$ of the pixels with forest mortality are predicted as changed.
We notice that parts of the predicted changes resemble outlines of the water bodies in Figure \ref{fig:CG2-a-five}, which indicates that CAE detects changes in water levels.
There are also changes in the agricultural and settled land, primarily along the Tana river.
The translation artefacts from lake Polmak in Figure \ref{subfig:cm2-trans-A-OPT} are also marked as changes.
Compared to these phenomena, which result in high magnitude difference vectors, the death of the canopy layer of the fragmented \fte{} is a subtle change at this resolution level.
To detect it we need to train a classifier to look for it specifically.

\subsubsection{Target change detection with proposed method}
\begin{figure}[h!]
        \includegraphics[width=\linewidth]{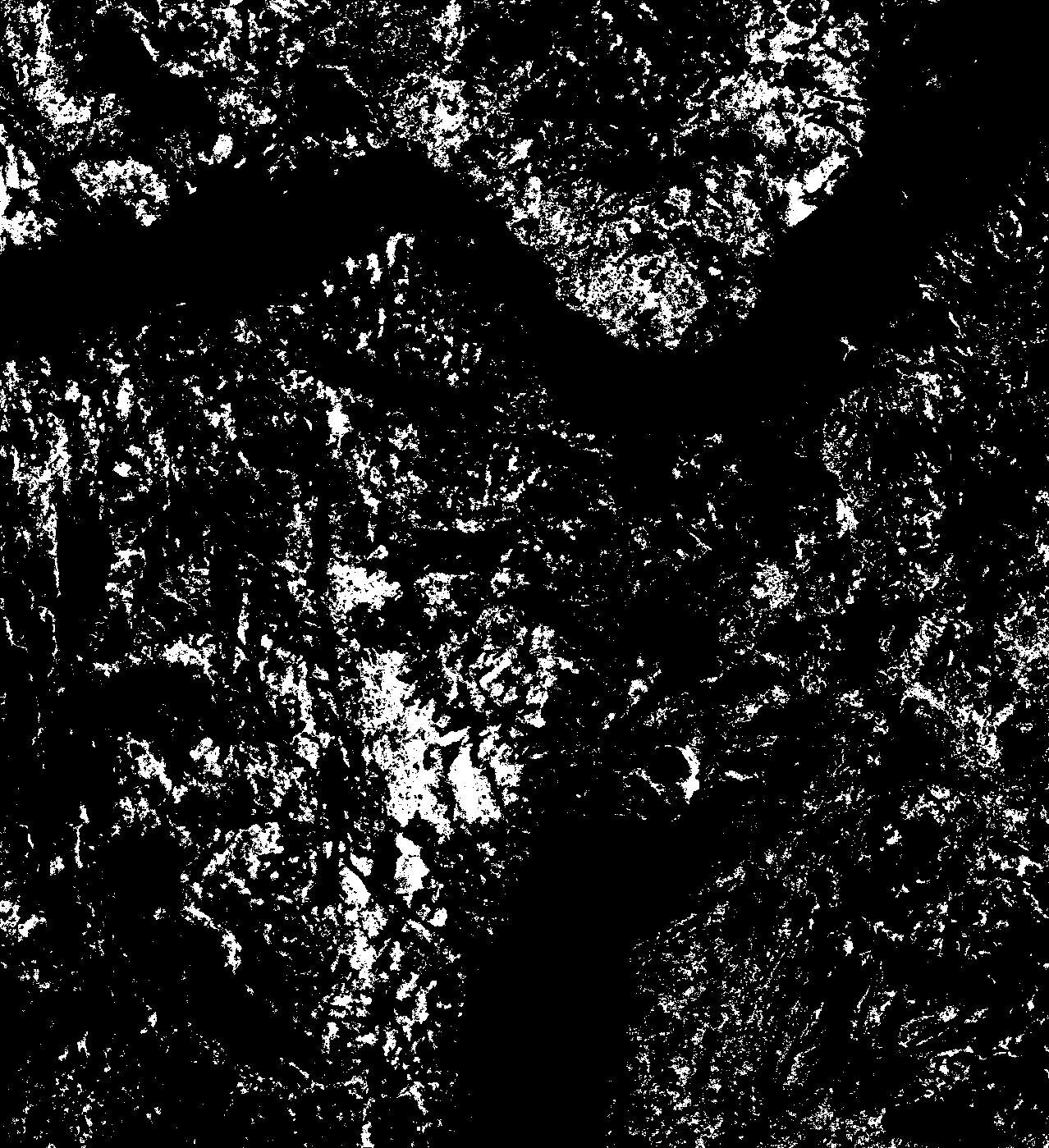}
        \caption[ Changemap ]
        {\small Predicted forest mortality areas shown in white using our approach.}
        \label{fig:LS2005-PGNLM_A}
\end{figure}

Figure \ref{fig:LS2005-PGNLM_A} shows the forest mortality map provided by our targeted change detection method.
The feature vectors in Eq.\ \eqref{eq:feature-vec} are obtained from the images and translations shown in Figure \ref{fig:CG2-a-five}.
The two-step OCC is then trained on 1536 feature vectors from the 15 polygons with known forest mortality.
The result indicates that large areas of forest have died following the outbreak.
Particularly the western side of lake Polmak, on the Finnish side of the border, has been heavily afflicted.
This is in line with the observations by \textcite{biuw2014long} that the regeneration of mountain birch stands appeared to have been severely hampered by the year round grazing on the Finnish side of the border.
Significant forest mortality is also detected north of the river.
We note the fragmented nature of the forest mortality map, which is natural given that the sparse nature of the \fte{}.
Contrary to the CAE result, there are no detections along the agricultural and settled land around the Tana river.

We do not have another map of the effects of the outbreak to evaluate against, so we cannot readily quantify the accuracy for the full forest mortality map.
By use of NDVI measurements from before and after the outbreak, in the same period as the input imagery, and correlating with the result in Figure \ref{fig:LS2005-PGNLM_A}, we find a noticeable NDVI decrease in the areas of classified forest mortality.
For these areas, shown as white in Figure \ref{fig:LS2005-PGNLM_A}, the NDVI decreased by $0.145$ on average.
In the other areas, the difference was virtually zero.

We also evaluate our result on a systematic grid of $\SI{30}{\metre}\times \SI{30}{\metre}$ cells aligned with transects examined during field work in 2017 \parencite{agersborg2021guided}.
The cells were classified by an expert into four classes based on a study of aerial photographs.
Three of the classes were related to forest mortality, as they describe the canopy state as "live" (33 cells), "dead" (44 cells), and "damaged" (30 cells) \parencite{agersborg2021guided}.
All cells with a canopy state class were from forest with live canopy in the 2005 aerial image, with the canopy state of the cell in the 2015 image determining if it was classified as live, damaged or dead.
The last class, denoted "other" (66 cells), is vegetation without canopy cover.
For each grid cell we find the pixels in the forest mortality map in Figure \ref{fig:LS2005-PGNLM_A} that intersects the geographical coordinates defining the bounding box of that cell.
Since the mortality map uses a $\SI{10}{\metre}\times \SI{10}{\metre}$ pixel spacing, each grid cell should correspond to $3\times3=9$ pixels, though a few will be larger, as we choose to include partially covered pixels.
Note that the grid is not aligned north-south and has not been co-registered to the satellite imagery.
Hence, there may be some inaccuracies due to misalignment, especially since the size of the cells are the same as the resolution as the LS-5 pre-event image.

For the classes "live" and "other", we observe a low false alarm rate with $\hat{\text{TN}}_\text{live grid} = 98.7\%$ and $\hat{\text{TN}}_\text{other grid} = 96.9\%$ of the pixels correctly classified as no forest mortality.
As part of the work to create the polygons with forest mortality, 15 polygons live forest were also extracted.
These were similar in shape (rectangular), size and grouping, and located relatively close to areas with forest mortality.
In total the 15 polygons with live forest consists of $2070$ pixels.
For the live polygons our approach performs well with $\hat{\text{TN}}_\text{live poly} = 97.2\%$, which is consistent with the result for the "live" class for the grid cells.
Note that the both the "live" and the "other" class are subsets of the negative superclass and are not necessarily accurate estimates for the true negative rate for the complete negative class.
Comparing the forest mortality map in Figure \ref{fig:LS2005-PGNLM_A} with the optical image in Figure \ref{subfig:cm2-org-LS2005} and the unsupervised CAE change detection result in Figure \ref{fig:cae-confmaps}, indicates that we avoid classifying changes in agricultural and settled areas as forest mortality, which is important for the true negative rate.

For the "dead" grid cells we observe a true positive rate of $\hat{\text{TP}}_\text{grid} = 66.4\%$.
Misalignment could be a contributing factor to the missed detections as 37 of 44 ($84.1\%$) of the cells do contain pixels classified as forest mortality.
The number of missed detections along with the low false alarm rate could also be an indication that our method is conservative in predicting forest mortality, though this was not seen for the training data where the true positive rate was $96.8\%$.
While accuracy scores for the training data always are of questionable validity, this indicates that our method does not have to "sacrifice" much accuracy on the positive training set to increase accuracy on the RNs during training.

It is not obvious if the grid cells with damaged canopy state should be considered as the positive  or negative class for evaluation. This state contains a mixture of dead stems and trees where parts of the canopy has re-sprouted.
$28.1\%$ of the pixels in the grid cells of this category were classified as forest mortality.
Arguments could be made for considering the damaged state as a part of the positive class, since there is a clear decrease in canopy cover.
However, for operational monitoring it could be more important to focus on areas where the forest has died completely, as it can be assumed that most of the forest will suffer canopy damage in a major outbreak, and our targeted change detection approach should be able to map only the former.
Nonetheless, that only $28.1\%$ of the pixels in the damaged class were labelled as forest mortality may be another indication that our prediction of forest mortality is on the conservative side.

\begin{figure*}[p]
        \centering
        \begin{subfigure}[b]{0.975\textwidth}
            \centering
            \includegraphics[width=\textwidth]{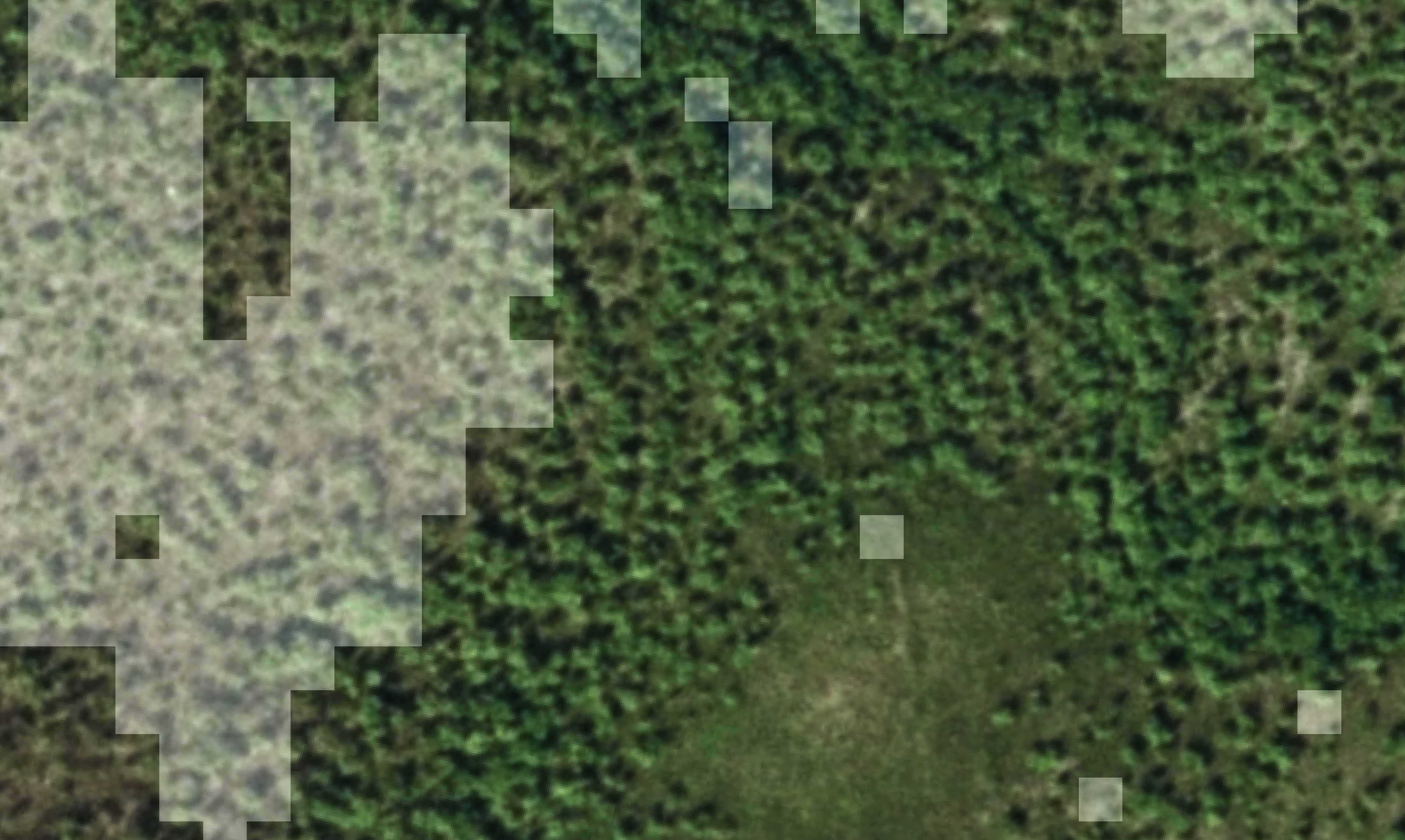}
            \caption[2005]%
            {{\small 2005}}
            \label{subfig:ds5-pre}
        \end{subfigure}
        \vskip\baselineskip
        \begin{subfigure}[b]{0.975\textwidth}
            \centering
            \includegraphics[width=\textwidth]{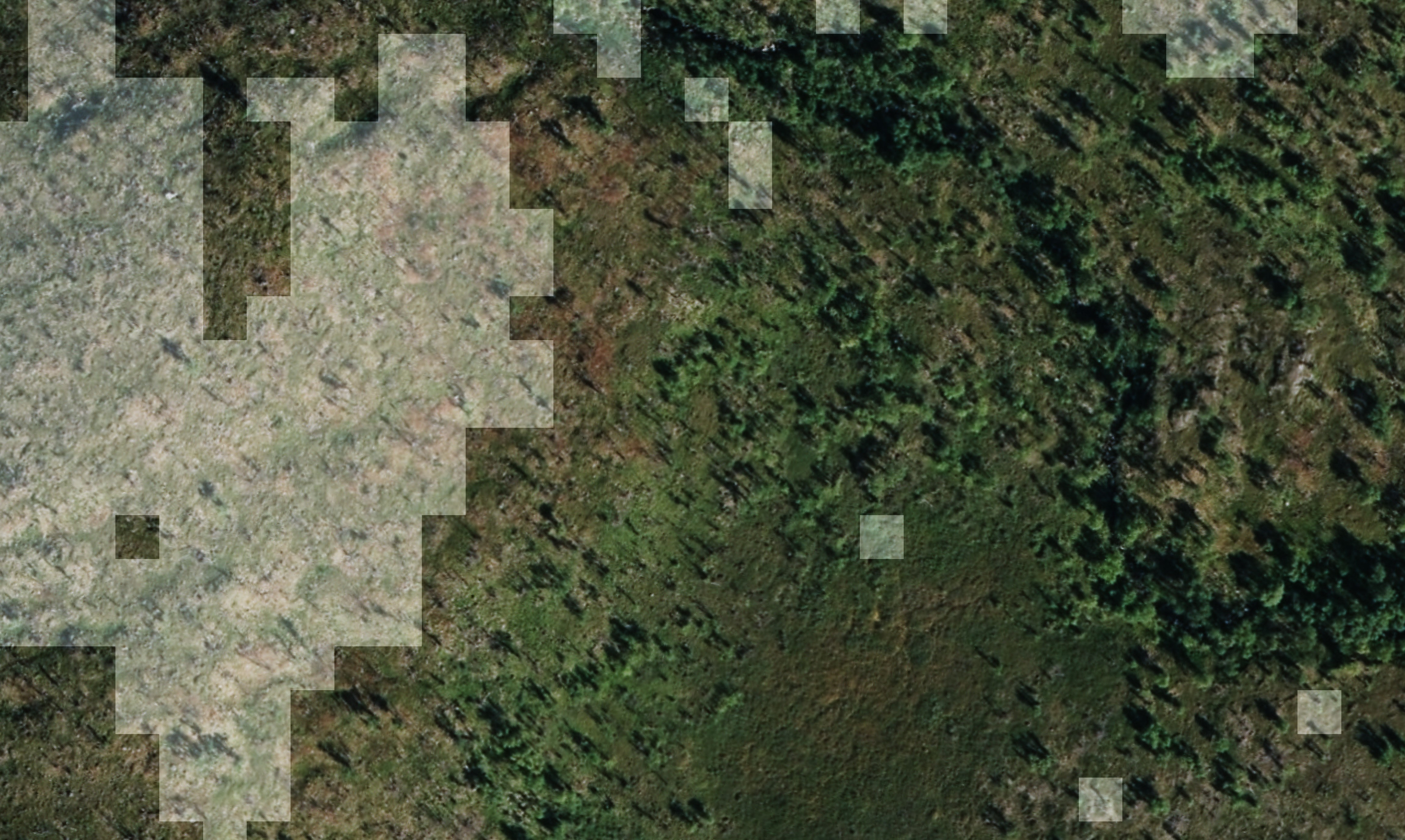}
            \caption[]%
            {{\small 2015}}
            \label{subfig:ds5-post}
        \end{subfigure}
        \caption[ Data and translation]
        {\small Forest mortality map overlain aerial photographs before and after the outbreak.}
        \label{fig:ds5}
\end{figure*}

To evaluate how well the forest mortality map corresponds to the high resolution aerial photographs,
we show the map as an overlay to the images from 2005 and 2015.
The mortality map is shown as a bright see-through layer in Figure \ref{fig:ds5}.
The image from 2005 is shown at the top of in Figure \ref{subfig:ds5-pre} with 2015 in Figure \ref{subfig:ds5-post}.
The latter image has a higher resolution of $\SI{0.25}{\metre}\times \SI{0.25}{\metre}$ compared to $\SI{0.50}{\metre}\times \SI{0.50}{\metre}$ for the 2005 image, and appears to be more easily interpretable.
The "pixelated" appearance of the mortality map is due to the $\SI{10}{\metre}\times \SI{10}{\metre}$ pixel spacing.
Most of the areas with forest mortality have been accurately detected, although the mapped area in the left half of Figure \ref{fig:ds5} could have included a larger area. There also appears to be some false alarms, with the three single $\SI{10}{\metre}\times \SI{10}{\metre}$ areas of the mortality map in the lower right half of the images in Figure \ref{fig:ds5} appear to have live canopy in the 2015 image.
Apart from that, our method appears to avoid misclassifying other landcover types as forest mortality.

\begin{figure*}[p]
        \centering
        \begin{subfigure}[b]{0.975\textwidth}
            \centering
            \includegraphics[width=\textwidth]{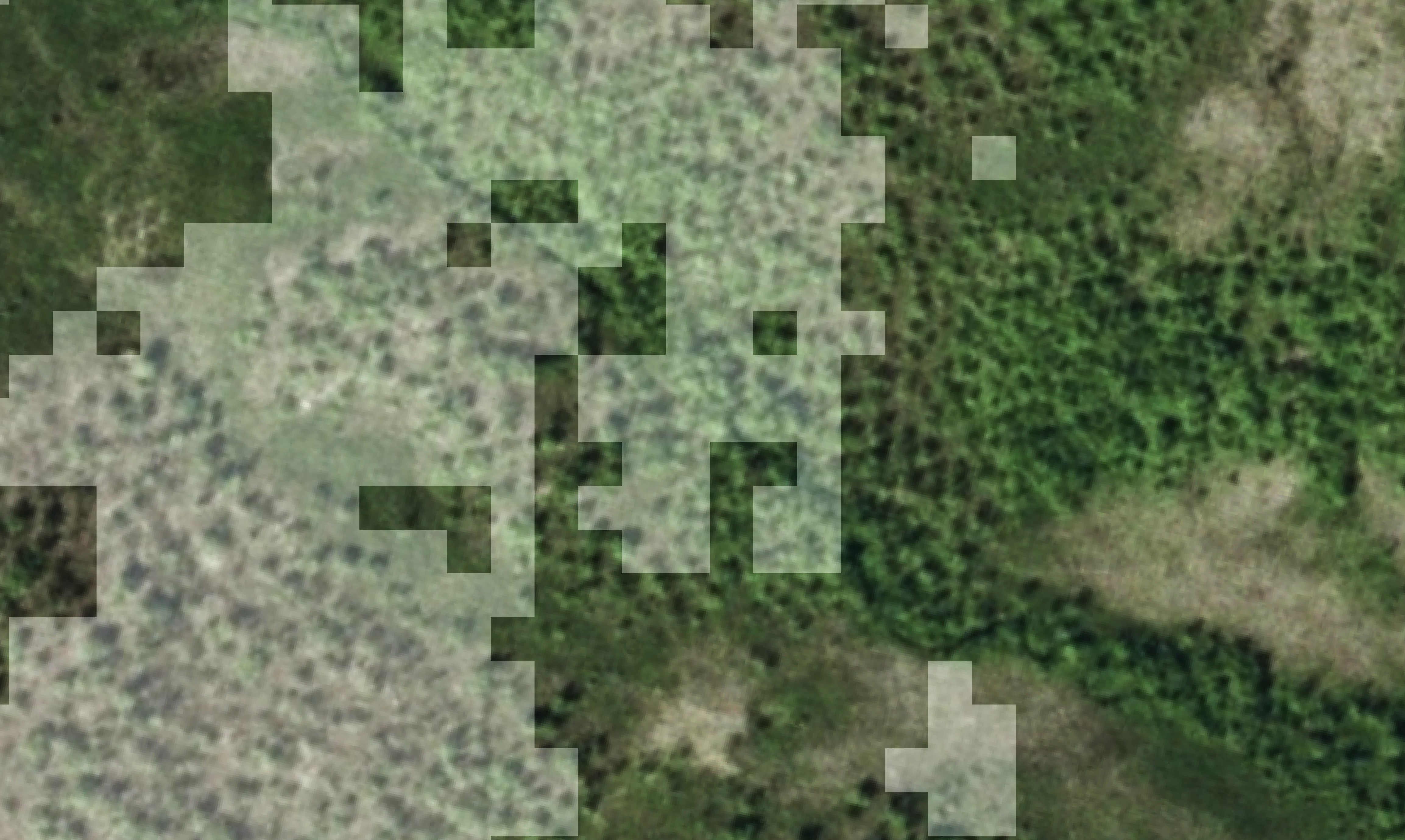}
            \caption[2005]%
            {{\small 2005}}
            \label{subfig:ds6-pre}
        \end{subfigure}
        \vskip\baselineskip
        \begin{subfigure}[b]{0.975\textwidth}
            \centering
            \includegraphics[width=\textwidth]{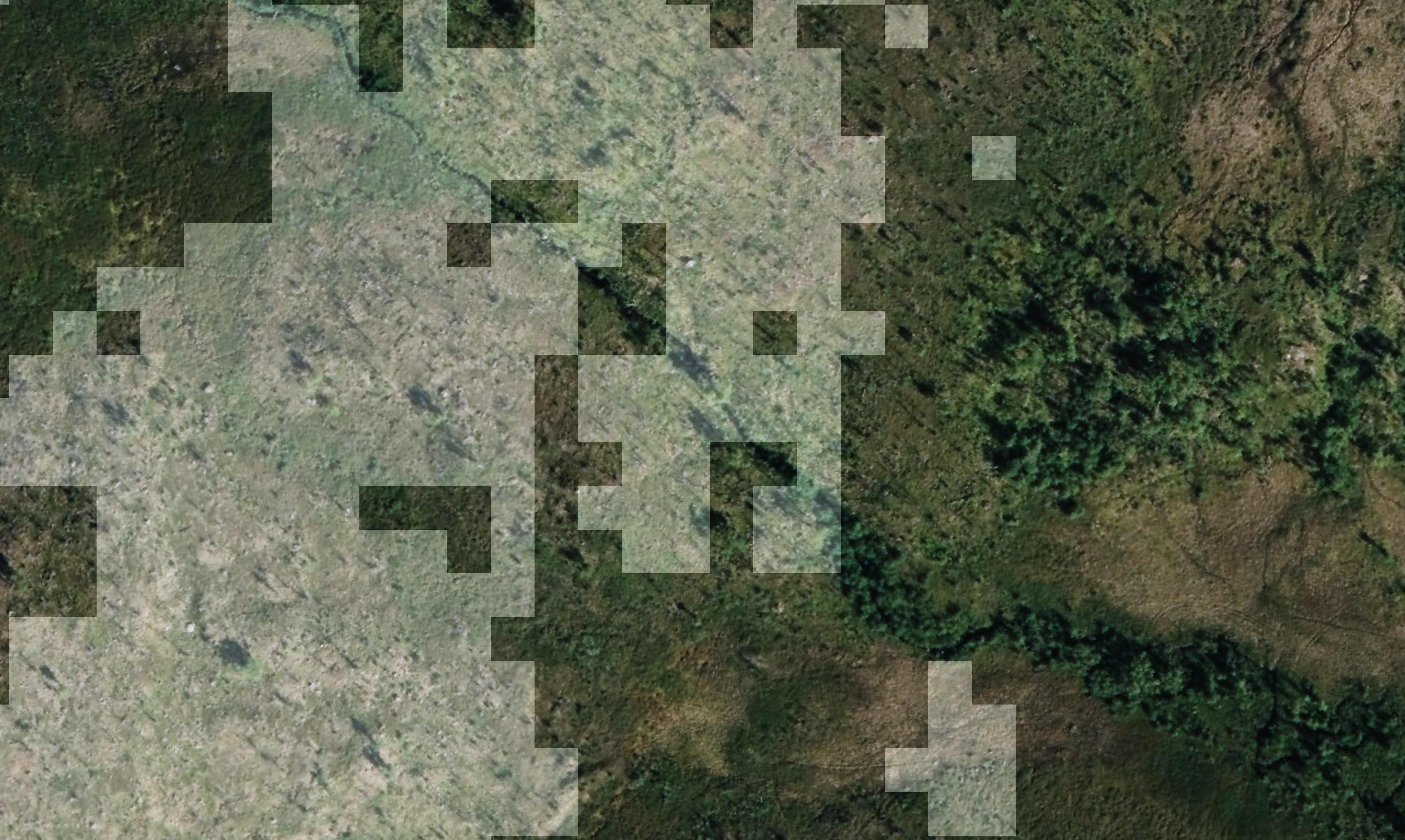}
            \caption[]%
            {{\small 2015}}
            \label{subfig:ds6-post}
        \end{subfigure}
        \caption[ Data and translation]
        {\small Forest mortality map overlain aerial images before and after the outbreak. }
        \label{fig:ds6}
\end{figure*}

We also note the challenge of accurately mapping forest mortality due to the sparse nature of the \fte{}, and the entangled pattern of live and dead trees.
Since the resulting map appears quite good, setting the pixel spacing of the mortality map to $\SI{10}{\metre}\times \SI{10}{\metre}$ to match the RS-2 resolution, seems warranted.
However, we should keep in mind that the resolution of LS-5 used as the pre-event image is $\SI{30}{\metre}\times \SI{30}{\metre}$, which will affect the accuracy of mortality map.
An example of the mortality map failing to delineate thin separations between classes is shown in Figure \ref{fig:ds6}.
Here there is a separation between areas where the trees have died that are included in the mortality map.
Similarly to the result in Figure \ref{fig:ds5}, the map shown in Figure \ref{fig:ds6} miss parts of the forest that has died which borders on the areas correctly mapped as dead.
It also appears that there is an area which is falsely classified as forest mortality in the bottom right part of Figure \ref{fig:ds6}.

\begin{figure*}[p]
        \centering
        \begin{subfigure}[b]{0.975\textwidth}
            \centering
            \includegraphics[width=\textwidth]{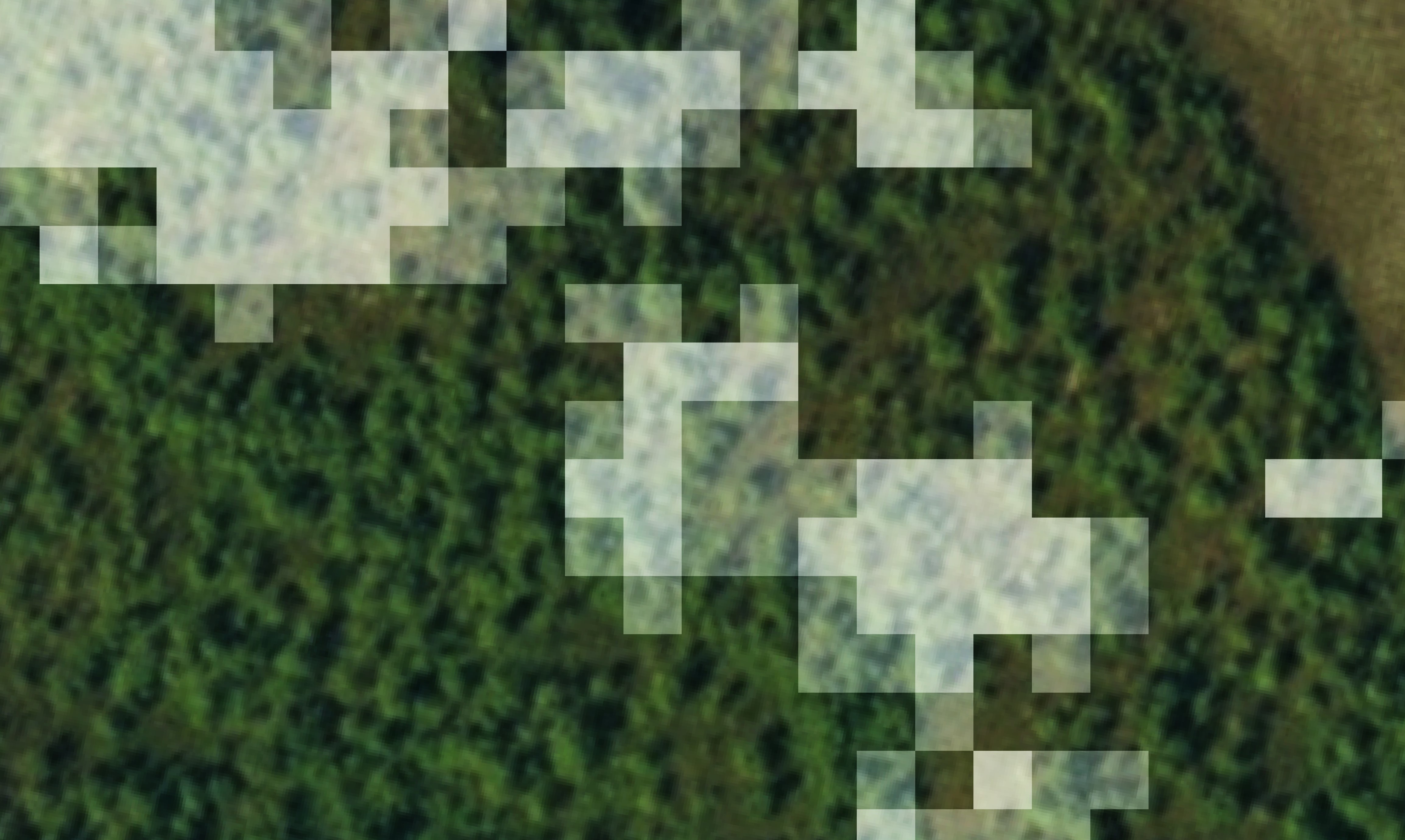}
            \caption[2005]%
            {{\small 2005}}
            \label{subfig:ds8-pre}
        \end{subfigure}
        \vskip\baselineskip
        \begin{subfigure}[b]{0.975\textwidth}
            \centering
            \includegraphics[width=\textwidth]{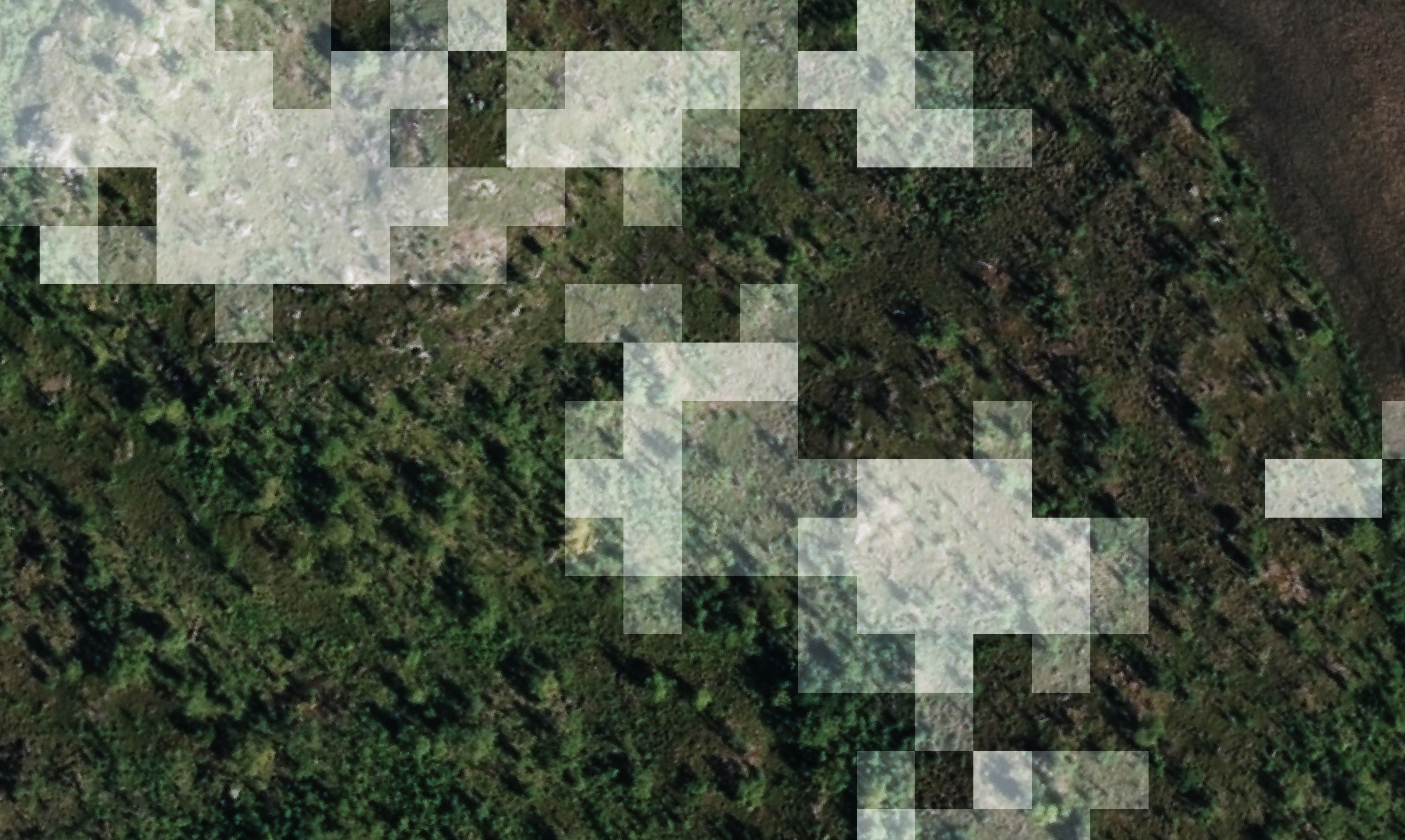}
            \caption[]%
            {{\small 2015}}
            \label{subfig:ds8-post}
        \end{subfigure}
        \caption[ Data and translation]
        {\small Two forest mortality maps overlain aerial images before and after the outbreak. The brightest overlay shows areas classified as forest mortality with both threshold, while the more transparent overlay are classifications only made with $t=0.3$.}
        \label{fig:ds8double}
\end{figure*}

The results in Figures \ref{fig:ds5} and \ref{fig:ds6}, and the numerical evaluations on the grid cells and live polygons, indicate that mortality map is conservative.
One way of easily increasing the number of pixels predicted as forest mortality is to adjust the ensemble voting done in the second step of the OCC method, where five MLPs classify the input data.
Reducing the number of MLPs required to predict the positive class would therefore increase the prediction of forest mortality.
This is equivalent to considering the output of the ensemble as the average number of predictors which predict the positive class as a number between $0$ and $1$ that is thresholded to give the final binary output map.
Then, reducing the votes necessary is the same as reducing this threshold $t$.
By reducing the number of votes required from three (majority) to two we obtain a higher true positive rate of $\hat{\text{TP}}_\text{grid} = 76.6\%$.
In terms of threshold, this corresponds to lowering the threshold from $t=0.5$ to $t=0.3$.
However, there is a decrease in the true negative rate, with $\hat{\text{TN}}_\text{live poly} = 94.6\%$, $\hat{\text{TN}}_\text{live grid} = 97.5\%$, and $\hat{\text{TN}}_\text{other grid} = 94.6\%$.
There was also an increase in the number of pixels from the "damaged" state classified as forest mortality to $33.7\%$.
When evaluating this result on the high resolution aerial photographs, we see that more forest mortality is correctly detected, but at the cost of a higher number of false alarms.
An example of this is shown in Figure \ref{fig:ds8double}, where the result of our method with $t=0.5$ to $t=0.3$ are both overlain the photographs from before and after the outbreak.
The brightest overlay areas are predicted as forest mortality with both thresholds, while the more see-through overlay corresponds to the prediction from $t=0.3$ alone.
For the area in the upper right corner, the $t=0.3$ appears better, while the difference in prediction at the bottom right of Figure \ref{fig:ds8double} appears to be mostly false alarms.

\section{Conclusions and future work}
\label{sec:conclusion}
In this work we have presented a method for detecting forest mortality from a pair of medium resolution heterogeneous remote sensing images to map the effect of a geometrid moth outbreak.
This is a challenging problem, and we were unable to achieve it using the unsupervised CAE results alone, since the phenomenon of interest has a weak signature compared to other changes.
By utilising the CAE for change aware image-to-image translation, we obtained multitemporal difference vectors despite the heterogeneity of the input images.
When combined with the original image features, a semi-supervised one-class classifier is able to learn to map the changes of interest from a very limited set of training data consisting of less than $0.1\%$ of the image pixels of these extended feature vectors.
The ablation study shows that all the different components of our targeted change detection approach contributes to the final output, and we achieve
good results for benchmark datasets, despite not being intended as a general change detection method.

The evaluation for our AOI indicates that we achieve a low false alarm rate, but that the predicted forest mortality map may be a bit conservative.
It is possible to increase the true positive rate at the expense of more false alarms by adjusting a threshold in the ensemble voting done in the second step of the OCC method.
However, we prefer a low false alarm rate to a higher number of detections in the trade-off, for instance in case the map should be used to determine new areas for field work to study the effect of the outbreak.
Future work should seek to assess performance on datasets with complete ground truth data available.
This should preferably be done on datasets suitable for targeted change detection, where changes unrelated to the phenomenon of interest are included in the negative class.

Our approach expands the potential for detecting the extent of changes that we know have occurred at one or more locations by using whatever satellite imagery available from before and after the event as long as it can be co-registered.
This allows us to map a phenomenon of interest over large areas.
It does not require a dense time series of data, in the same manner as NDVI-based approaches, which can be problematic for our AOI given the high cloud cover percentage.
The modular nature of our approach means that components can be replaced if the particular dataset warrants it.
This could also be investigated in future work, and our ablation study hints at some interesting directions.

\section*{Disclosure statement}
The authors declare no conflicts of interest.

\section*{Funding}
This work was supported in by the COAT Tools project (www.coat.no/en/Education/COAT-Tools) by an Interdisciplinary Project Grant from the Developing the High North Strategic Programme of UiT The Arctic University of Norway; and by the Research Council of Norway under Grant 301922.

\section*{ORCID}
Jørgen~A.~Agersborg \orcidlink{0000-0002-1526-6659} \url{https://orcid.org/0000-0002-1526-6659} \\
Luigi~T.~Luppino \orcidlink{0000-0002-1288-817X} \url{https://orcid.org/0000-0002-1288-817X} \\
Stian~Normann~Anfinsen \orcidlink{0000-0002-3758-4295} \url{https://orcid.org/0000-0002-3758-4295} \\
Jane~Uhd~Jepsen \orcidlink{0000-0003-1517-1569} \url{https://orcid.org/0000-0003-1517-1569}

\appendix

\section*{Appendix}

\subsection{Code-aligned autoencoders}
\label{sec:appendix-cae}

As the name implies, the CAE algorithm uses an autoencoder architecture that learns a pair of convolutional neural networks, the encoder and the decoder, for each of the images.
The domain-specific encoders are trained to encode their respective input images into a code representation, while the decoders are trained to reconstruct the input images with high fidelity from these codes.
That is, if we denote the encoder associated with the pre- and post-event images as $\preencoder$ and $\postencoder$, respectively, these can be used to obtain encoded representations of $\preimage$ and $\postimage$ as $\encodepre{\preimage} = \codepre$ and $\encodepost{\postimage} = \codepost$.
The corresponding decoders, $\predecoder$ and $\postdecoder$, then reconstruct the encoded input as:
\begin{align}
    \decodepre{\codepre} = \decodepre{\encodepre{\preimage}} & = \Tilde{\preimage} \approx \preimage  \label{eq:reconstruct-domain1} \\
    \decodepost{\codepost} = \decodepost{\encodepost{\postimage}} & = \Tilde{\postimage} \approx \postimage \label{eq:reconstruct-domain2}
\end{align}
where the reconstructed original pre- and post-event images, $\Tilde{\preimage}$ and $\Tilde{\postimage}$, should be approximately equal to the corresponding original images.
This objective is formulated as a loss function, the reconstruction loss, which is an inherent part of all autoencoders.
For CAE the reconstruction loss is one of four terms in the total loss function used to train the full network.

\begin{figure}[h!]
  \centering
  \includegraphics[width=1.0\linewidth, keepaspectratio]{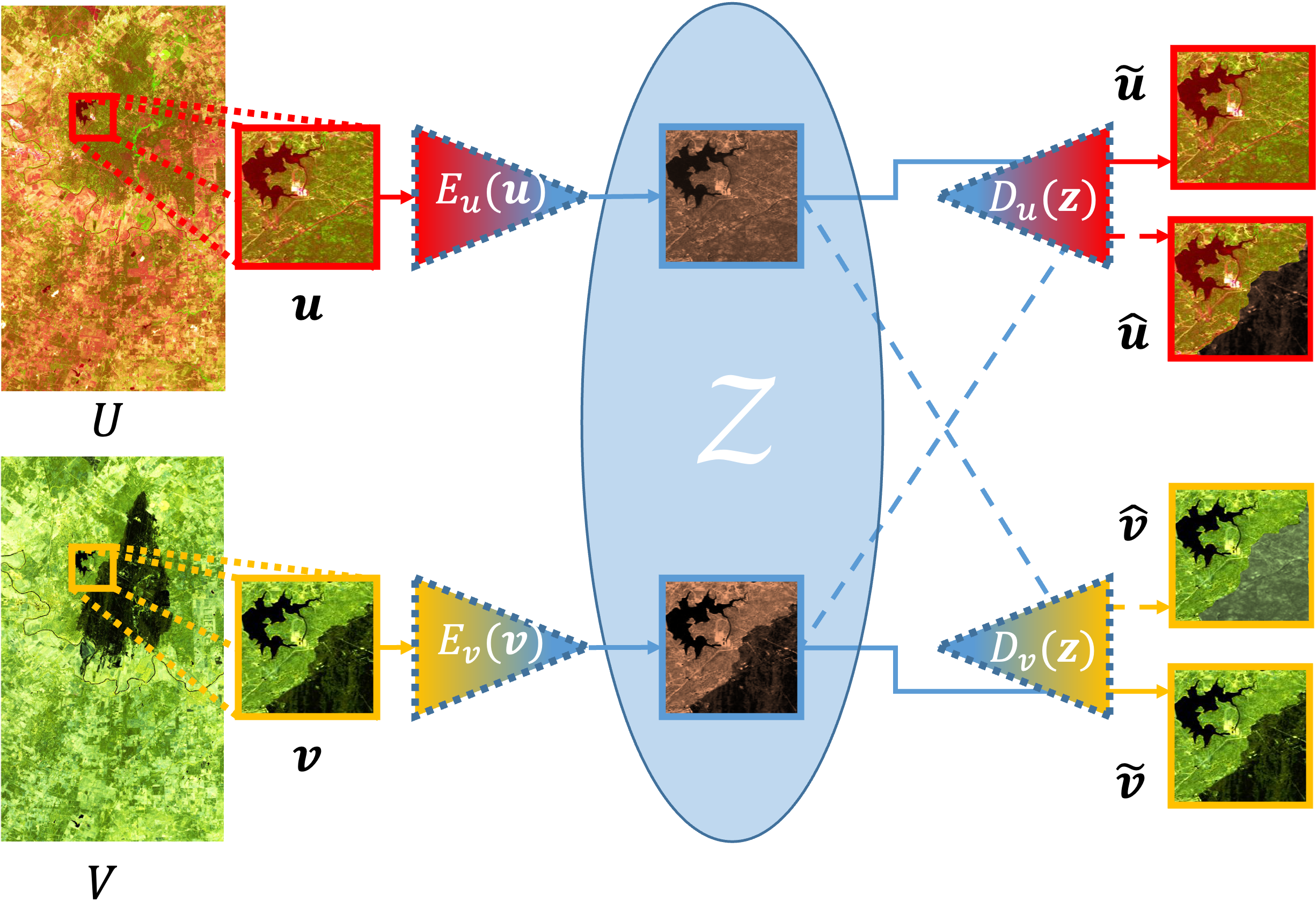}
\caption{Illustration of code-aligned autoencoder network showing the translation of patches $\prevec$ from $\preimage$ and $\postvec$ from $\postimage$.}
\label{fig:cae-illustration}
\end{figure}

In general, the code layer representation of two separately trained autoencoders are not similar.
By aligning the code spaces one can obtain a translation between domains by using the decoder from one domain on the coded representation of the other domain.
That is
\begin{align}
    \decodepre{\codepost} = \decodepre{\encodepost{\postimage}} & = \hatpreimage  \label{eq:trans-to-domain1} \\
    \decodepost{\codepre} = \decodepost{\encodepre{\preimage}} & = \hatpostimage \label{eq:trans-to-domain2}
\end{align}
where the encoders and decoders are the same as for Equations \eqref{eq:reconstruct-domain1} and \eqref{eq:reconstruct-domain2}, and $\hatpreimage$ and $\hatpostimage$ are the pre- and post-event images translated to the other domain. Figure \ref{fig:cae-illustration}, adapted from \textcite{luigi2020paper3}, illustrates the network, showing the result of encoding and decoding a pair of coregistered image patches from the Texas dataset.

CAE enforces alignment of the code layers of the two autoencoders by adding a loss terms that ensures their alignment in both distribution and location of land covers \parencite{luigi2020paper3}.
This code correlation loss is a novel feature of CAE, and enforces that $\codepre$ should be similar to $\codepost$ \parencite{luigi2020paper3}.
The similarity is based on a cross-modal distance between training patches in the input domains.
This allows pixels that have changed to be distinguished from those who have not, and the loss term seeks to preserve these relationships in the code layer.
Contrary to the other loss functions, which are used to train the both encoders and decoders, the code correlation loss is only used for the encoders.

A cycle-consistency loss enforces that data translated from one domain to the other, and then back again, should be identical to the input.
In a sense it is similar to the reconstruction loss, except that the cycle-consistency loss involves all encoders and decoders of the network.

The final loss term requires that the translations $\hatpreimage$ and $\hatpostimage$ in Equations \eqref{eq:trans-to-domain1} and \eqref{eq:trans-to-domain2} should be similar to the data in the original domain $\preimage$ and $\postimage$, except for pixels where changes have occurred.
If there is a significant chance that a change has occurred for a particular pixel, its contribution to this loss term is strongly suppressed, whereas pixels of $\hatpreimage$ ($\hatpostimage$) from likely unchanged areas should be close to $\preimage$ ($\postimage$).
To distinguish between changed and unchanged areas, this loss term includes a weighting factor updated iteratively during training. This is based on preliminary change detection results obtained with the image translations at the current stage of training.
Note that while the final change-detection result of CAE is a binary change map, the weighting factor is a continuous variable between zero and one where lower values indicates where there is high probability of change.

We have made some adaptations to the CAE related to network training, with the aim of improving the visual quality and detail preservation of the translations, as briefly summarised:
\begin{itemize}
    \item The patch size used for training is reduced from $100 \times 100$ pixels to $20 \times 20$, so the code correlation loss is calculated for all pixel pairs in the input patches.
    In the original implementation the cross-modal distance between training patches were based on a $20 \times 20$ pixels excerpt from the centre of the full $100 \times 100$ training patch due to memory constraints \parencite{luigi2020thesis}.
    By reducing its size, the full training patch is used for the code correlation loss to better align the two domains.
    \item The number of patches per training batch is increased from $10$ to $20$ and the number of batches per epoch from $10$ to $600$ to compensate for the lower number of pixels seen during training due to the reduced patch size.
    A $100 \times 100$ patch contains $25$ times as many pixels as a $20 \times 20$ one, and it is therefore necessary to increase the batch size and the number of epochs compared to \textcite{luigi2020thesis}.
    \item The preliminary evaluation of the difference image is changed from using the reconstructed versions ($\Tilde{\preimage}$ and $\Tilde{\postimage}$) to using the originals in the weighted translation loss to better preserve details in the translations.
\end{itemize}
In our experience, these modifications improve the visual quality of the translations, and through more meaningful difference images also the accuracy of the final classification.

\subsection{One-class classification}
\label{app:occ}

OCC is framed as a binary classification problem, where the positive class of interest has label $\classlabel=1$ and the negative class, $\classlabel=0$, is usually defined as the complement of the positive class \parencite{li2020one}.
The full dataset $\dataset$ is typically divided into $\posiset$, the set of labelled positive samples, and an unlabelled set $\unlaset$ (also called mixed set) that consists of data from both the positive and the negative class.

Text classification and document retrieval are applications where OCC has seen much use and several OCC methods that have been developed are customised for the text domain.
In the taxonomy of OCC techniques proposed by \textcite{khan2014one}, the applications were divided into two categories: "text/document classification" and "other applications".
The terms single-class classification (SCC) \parencite{yu2005single}, partially supervised classification \parencite{liu2002partially}, and others (see \textcite{khan2014one} for a brief summary) have been used for one-class classification problem.
OCC is also related to positive and unlabelled learning (PUL), and the distinction between the two is somewhat blurry.
Just as in the OCC setting, PUL assumes that a labelled positive training set and an unlabelled set that contains mixed samples from both the positive and negative classes are available.
Many PUL methods are based on estimating dataset attributes, such as the labelling frequency and class prior probability of the positive class, and thus needs to make assumptions about how the labelled positive samples were generated \parencite{bekker2020learning}.
Some methods assume that the labelled positive samples were selected completely at random (SCAR).
PUL also considers two different settings: the single-training-set scenario, where the positive and unlabelled samples are from the same dataset, and the case-control scenario, where they are from different datasets.
We choose to draw the distinction between OCC and PUL by saying that the latter makes assumptions about how the positive samples were labelled, or that it is designed to work in the case-control scenario, or both.
Further, unlike PUL, OCC also opens up for the possibility that some labelled negative data may be available, although not well enough sampled or statistically representative enough to build a traditional binary classifier.
Note that the PUL acronym has also been used by \textcite{elkan-noto2008learning} about a learning algorithm using the SCAR assumption and a specially held-out validation set to estimate the probability of a positive sample being labelled.

The SCAR assumption does not hold for our application with the available ground reference data from our AOI.
The weaker "selected at random" condition, which assumes that labelled examples "are a biased sample from the positive distribution, where the bias completely depends on the attributes" \parencite{bekker2020learning}, also does not hold.
Our ground reference data is selected systematically from a limited area, and the sample selection bias is related to the attributes of the feature vectors due to some spatial correlation in type of background vegetation, soil conditions, tree densities and mortality rates, etc.
However, this bias is not completely dependent on the attributes of the feature vectors.
Therefore, according to our definition above, we use the term OCC and not PUL in this work, although we acknowledge that the distinction between the terms is not well established.

There are many different OCC methods to choose from, and the choice should naturally depend on the application.
The OCC taxonomy by \textcite{khan2014one} divides the methodology into "one-class support vector machine" (OCSVM) and "non-OCSVM".
However, this taxonomy \parencite{khan2014one} is focused on the text domain and only mentions in passing other application domains without summarising which OCC methods they use.
A more useful taxonomy, at least when it comes to guiding our choice of methods, is provided in the review article by \textcite{bekker2020learning}, which lists three different categories of PUL methods: two-step techniques, biased learning, and class prior incorporation.
The latter two invoke the SCAR assumption, and are thus not applicable in our case.
We therefore end up using the two-step technique, which was previously reviewed.

OCC has been used to extract a particular land cover type from remote sensing images. It has also been applied for targeted change detection. \textcite{camps2008kernel} tested a one-class kernel support vector domain descriptor (SVDD) method for heterogeneous change detection of transitions between urban and non-urban landcover on an optical-SAR image pair. SVDD is similar to one-class SVM (OCSVM), but uses a hypersphere instead of a hyperplane for separation \parencite{munoz2010semisupervised}. It can be shown that with normalised data and isotropic kernels, the two methods give the same results \parencite{munoz2010semisupervised}. The SVDD compared favourably to a simple single hidden layer multilayer perceptron (MLP) and other SVM classifiers trained on both changed and unchanged pixels \parencite{camps2008kernel}.

\textcite{li2010land} used OCSVM with a radial basis function (RBF) kernel to detect changes in two bitemporal pairs with features derived from Landsat-5 Thematic Mapper (TM) images.
In the more complicated experiment of urban expansion, OCSVM was used separately for each of the different change type (water-urban, soil-urban, vegetation-urban) \parencite{li2010land}, thus requiring three separate classification procedures and three different labelled training datasets.
OCSVM compared favourably to post-classification comparison (PCC) in tests performed on balanced samples from both datasets.

Two modified SVM methods were tested on four remote sensing problems by \textcite{munoz2010semisupervised}, one of which was a homogeneous change detection problem.
The two methods considered were variations of OCSVM and biased SVM \parencite{liu2003building} adjusted to better utilise the unlabelled data.
OCSVM was modified to include unlabelled data to adjust the SVM kernel, while the modification of biased SVM was an adjustment to the cost function \parencite{munoz2010semisupervised}.
However, only a small subset of the unlabelled samples were used due to the computational cost of inverting the kernel matrix that increases exponentially with the number of samples \parencite{munoz2010semisupervised}.

An OCC method based on sparse representation of the features was tested for bitemporal change detection of a flood in a homogeneous multispectral dataset and compared with an RBF-kernel OCSVM by \textcite{ran2016change}.
\textcite{ran2018kernel} reported results using kernelised versions of the sparse representation classifiers.

\textcite{ye2016targeted} presented an OCC-based method for targeted change detection combining SVDD with results from change vector analysis (CVA) and PCC. However, due to difficulties in finding tight boundaries for the target cluster in the feature space, the changed class was subdivided into several subclasses depending on the spectral signature, each dependent on a balanced training dataset \parencite{ye2016targeted}. The approach was tested on homogeneous image pairs, and good results were reported on selected balanced training and test datasets.

\textcite{jian2021gan} used generative adversarial networks (GANs) to perform OCC change detection.
It was based on "spatial-spectral" features extracted from a stack of three homogeneous RGB remote sensing images, where the change occurred in the latest image of the stack.
Contrary to the other methods discussed, the training dataset was created from unchanged data from the first two images of the stack.
Two GANs generate samples from a "change data" distribution, and a discriminator was trained to separate unchanged from generated change samples \parencite{jian2021gan}.
This discriminator was then finally used to detect changes in the spatial-spectral features generated from the second and third images in the stack where the change occurred.
The method performed comparably to deep learning based unsupervised change detection methods and other OCC methods when these, contrary to the normal setting of labelled positive data, were trained with data from the unchanged negative class.
It should be noted that while the concept of using only unchanged data for OCC-based change detection is intriguing, it requires additional unchanged images and is unable to perform targeted change detection.

\printbibliography

@Article{volpi2015spectral,
  author    = {Volpi, Michele and Camps-Valls, Gustau and Tuia, Devis},
  title     = {Spectral alignment of multi-temporal cross-sensor images with automated kernel canonical correlation analysis},
  pages     = {50--63},
  volume    = {107},
  journal   = {ISPRS J.\ Photogram.\ Remote Sens.},
  publisher = {Elsevier},
  year      = {2015},
}

@Article{bae2021tracking,
  author    = {Bae, Soyeon and M{\"u}ller, J{\"o}rg and F{\"o}rster, Bernhard and Hilmers, Torben and Hochrein, Sophia and Jacobs, Martin and Leroy, Benjamin ML and Pretzsch, Hans and Weisser, Wolfgang W and Mitesser, Oliver},
  title     = {Tracking the temporal dynamics of insect defoliation by high-resolution radar satellite data},
  journal   = {Methods in Ecology and Evolution},
  publisher = {Wiley Online Library},
  year      = {2021},
}

@InProceedings{li2003learning,
  author       = {Li, Xiaoli and Liu, Bing},
  booktitle    = {IJCAI},
  title        = {Learning to classify texts using positive and unlabeled data},
  number       = {2003},
  organization = {Citeseer},
  pages        = {587--592},
  volume       = {3},
  year         = {2003},
}

@InProceedings{liu2002partially,
  author    = {Liu, Bing and Lee, Wee Sun and Yu, Philip S and Li, Xiaoli},
  booktitle = {ICML},
  title     = {Partially supervised classification of text documents},
  number    = {485},
  pages     = {387--394},
  volume    = {2},
  year      = {2002},
}

@Article{jepsen2009monitoring,
  author    = {Jepsen, Jane Uhd and Hagen, Snorre B. and H{\o}gda, K. A. and Ims, Rolf A. and Karlsen, Stein Rune and T{\o}mmervik, Hans and Yoccoz, Nigel G.},
  title     = {Monitoring the spatio-temporal dynamics of geometrid moth outbreaks in birch forest using {MODIS}-{NDVI} data},
  number    = {9},
  pages     = {1939--1947},
  volume    = {113},
  journal   = {Remote Sens.\ Environ.},
  publisher = {Elsevier},
  year      = {2009},
}

@Article{scikit-learn,
  author    = {Pedregosa, F. and Varoquaux, G. and Gramfort, A. and Michel, V. and Thirion, B. and Grisel, O. and Blondel, M. and Prettenhofer, P. and Weiss, R. and Dubourg, V. and Vanderplas, J. and Passos, A. and Cournapeau, D. and Brucher, M. and Perrot, M. and Duchesnay, E.},
  title     = {Scikit-learn: Machine Learning in {P}ython},
  pages     = {2825--2830},
  volume    = {12},
  journal   = {Journal of Machine Learning Research},
  year      = {2011},
}

@Article{biuw2014long,
  author    = {Biuw, Martin and Jepsen, Jane U and Cohen, Juval and Ahonen, Saija H and Tejesvi, Mysore and Aikio, Sami and W{\"a}li, Piippa R and Vindstad, Ole Petter L and Markkola, Annamari and Niemel{\"a}, Pekka and Ims, Rolf A},
  title     = {Long-term impacts of contrasting management of large ungulates in the Arctic tundra-forest ecotone: ecosystem structure and climate feedback},
  number    = {5},
  pages     = {890--905},
  volume    = {17},
  journal   = {Ecosystems},
  publisher = {Springer},
  year      = {2014},
}

@PhdThesis{luigi2020thesis,
  author    = {Luppino, Luigi Tommaso},
  title     = {Unsupervised Change Detection in Heterogeneous Remote Sensing Imagery},
  school    = {UiT The Arctic University of Norway, Dept.\ of Physics and Technology},
  year      = {2020},
}

@Article{olsson2016development,
  author    = {Olsson, Per-Ola and Kantola, Tuula and Lyytik{\"a}inen-Saarenmaa, P{\"a}ivi and J{\"o}nsson, Anna Maria and Eklundh, Lars and others},
  title     = {Development of a method for monitoring of insect induced forest defoliation-limitation of {MODIS} data in {F}ennoscandian forest landscapes},
  journal   = {Silva Fennica},
  year      = {2016},
}

@Article{li2010positive,
  author    = {Li, Wenkai and Guo, Qinghua and Elkan, Charles},
  title     = {A positive and unlabeled learning algorithm for one-class classification of remote-sensing data},
  number    = {2},
  pages     = {717--725},
  volume    = {49},
  journal   = {IEEE Trans.\ Geosci.\ Remote Sens.},
  publisher = {IEEE},
  year      = {2010},
}

@Article{olsson2016near,
  author    = {Olsson, Per-Ola and Lindstr{\"o}m, Johan and Eklundh, Lars},
  title     = {Near real-time monitoring of insect induced defoliation in subalpine birch forests with {MODIS} derived {NDVI}},
  pages     = {42--53},
  volume    = {181},
  journal   = {Remote Sens.\ Environ.},
  publisher = {Elsevier},
  year      = {2016},
}

@Article{kingma2014adam,
  author    = {Kingma, Diederik P and Ba, Jimmy},
  title     = {Adam: A method for stochastic optimization},
  journal   = {arXiv preprint arXiv:1412.6980},
  year      = {2014},
}

@Article{bekker2020learning,
  author    = {Bekker, Jessa and Davis, Jesse},
  title     = {Learning from positive and unlabeled data: A survey},
  number    = {4},
  pages     = {719--760},
  volume    = {109},
  journal   = {Machine Learning},
  publisher = {Springer},
  year      = {2020},
}

@InProceedings{elkan-noto2008learning,
  author    = {Elkan, Charles and Noto, Keith},
  booktitle = {Proc.\ 14th ACM SIGKDD Int.\ Conf.\ on knowledge discovery and data mining},
  title     = {Learning classifiers from only positive and unlabeled data},
  pages     = {213--220},
  year      = {2008},
}

@InProceedings{liu2003building,
  author       = {Liu, Bing and Dai, Yang and Li, Xiaoli and Lee, Wee Sun and Yu, Philip S},
  booktitle    = {Third IEEE Int.\ Conf.\ on Data Mining},
  title        = {Building text classifiers using positive and unlabeled examples},
  organization = {IEEE},
  pages        = {179--186},
  year         = {2003},
}

@Article{khan2014one,
  author    = {Khan, Shehroz S and Madden, Michael G},
  title     = {One-class classification: taxonomy of study and review of techniques},
  number    = {3},
  pages     = {345--374},
  volume    = {29},
  journal   = {The Knowledge Engineering Review},
  publisher = {Cambridge University Press},
  year      = {2014},
}

@Article{perbet2019near,
  author    = {Perbet, Pauline and Fortin, Michelle and Ville, Anouk and B{\'e}land, Martin},
  title     = {Near real-time deforestation detection in {Malaysia} and {Indonesia} using change vector analysis with three sensors},
  number    = {19},
  pages     = {7439--7458},
  volume    = {40},
  journal   = {Int.\ J.\ Remote Sens.},
  publisher = {Taylor \& Francis},
  year      = {2019},
}

@Article{touati2019multimodal,
  author    = {Touati, Redha and Mignotte, Max and Dahmane, Mohamed},
  title     = {Multimodal change detection in remote sensing images using an unsupervised pixel pairwise-based {Markov} random field model},
  pages     = {757--767},
  volume    = {29},
  journal   = {IEEE Trans.\ Image Process.},
  publisher = {IEEE},
  year      = {2019},
}

@TechReport{coat2021tundra,
  author      = {Pedersen, {\AA}shild {\O}nvik and Jepsen, Jane Uhd and Paulsen, Ingrid Marie Garfelt and Fuglei, Eva and Mosbacher, Jesper B and Ravolainen, Virve and Yoccoz, Nigel G and {\O}seth, Ellen and B{\"o}hner, H and Br{\aa}then, K A and Ehrich, D and Henden, J-A and Isaksen, K and Jakobsson, S and Madsen, J and Soininen, Eeva and Stien, A and Tombre, I and Tveraa, T and Tveito, O E and Vindstad, Ole Petter L and Ims, Rolf A},
  institution = {Norsk Polarinstitutt},
  title       = {Norwegian Arctic tundra: a panel-based assessment of ecosystem condition},
  year        = {2021},
}

@Misc{COATscienceplan,
  author       = {Ims, Rolf A. and Jepsen, Jane Uhd and Stien, Audun and Yoccoz, Nigel G.},
  title        = {Science Plan for {COAT}: {Climate-ecological} {Observatory} for {Arctic} {Tundra}},
  howpublished = {Fram Centre Report Series 1},
  note         = {{Fram} {Centre}, {Troms{\o}}},
  address      = {{Troms{\o}}, {Norway}},
  publisher    = {Fram Centre},
  year         = {2013},
}

@Article{agersborg2021guided,
  author    = {Agersborg, J{\o}rgen A and Anfinsen, Stian Normann and Jepsen, Jane Uhd},
  title     = {Guided Nonlocal Means Estimation of Polarimetric Covariance for Canopy State Classification},
  doi       = {10.1109/TGRS.2021.3090831},
  journal   = {IEEE Trans.\ Geosci.\ Remote Sens.)},
  volume={60},
  number={5208417},
  pages={1-17},
  year      = {2021},
}

@Article{gao2020remote,
  author    = {Gao, Yan and Skutsch, Margaret and Paneque-G{\'a}lvez, Jaime and Ghilardi, Adrian},
  title     = {Remote sensing of forest degradation: a review},
  number    = {10},
  pages     = {103001},
  volume    = {15},
  journal   = {Environmental Res.\ Lett.},
  publisher = {IOP Publishing},
  year      = {2020},
}

@Article{ye2016targeted,
  author    = {Ye, Su and Chen, Dongmei and Yu, Jie},
  title     = {A targeted change-detection procedure by combining change vector analysis and post-classification approach},
  pages     = {115--124},
  volume    = {114},
  journal   = {ISPRS J.\ Photogram.\ Remote Sens.},
  publisher = {Elsevier},
  year      = {2016},
}

@Article{dempster1977em,
  author    = {Dempster, Arthur P and Laird, Nan M and Rubin, Donald B},
  title     = {Maximum likelihood from incomplete data via the {EM} algorithm},
  number    = {1},
  pages     = {1--22},
  volume    = {39},
  journal   = {J.\ Royal Statist.\ Soc.: Ser.\ B (Methodological)},
  publisher = {Wiley Online Library},
  year      = {1977},
}

@Article{li2010land,
  author    = {Li, Peijun and Xu, Haiqing},
  title     = {Land-cover change detection using one-class support vector machine},
  number    = {3},
  pages     = {255--263},
  volume    = {76},
  journal   = {Photogrammetric Engineering \& Remote Sensing},
  publisher = {American Society for Photogrammetry and Remote Sensing},
  year      = {2010},
}

@Article{yu2005single,
  author    = {Yu, Hwanjo},
  title     = {Single-class classification with mapping convergence},
  number    = {1-3},
  pages     = {49--69},
  volume    = {61},
  journal   = {Machine Learning},
  publisher = {Springer},
  year      = {2005},
}

@Misc{hsu2003practical,
  author    = {Hsu, Chih-Wei and Chang, Chih-Chung and Lin, Chih-Jen},
  title     = {A practical guide to support vector classification},
  year      = {2003},
}

@Article{sun2021nonlocal,
  author    = {Sun, Yuli and Lei, Lin and Li, Xiao and Sun, Hao and Kuang, Gangyao},
  title     = {Nonlocal patch similarity based heterogeneous remote sensing change detection},
  pages     = {107598},
  volume    = {109},
  journal   = {Pattern Recognition},
  publisher = {Elsevier},
  year      = {2021},
}

@Article{ran2016change,
  author    = {Ran, Qiong and Zhang, Mengmeng and Li, Wei and Du, Qian},
  title     = {Change detection with one-class sparse representation classifier},
  number    = {4},
  pages     = {042006},
  volume    = {10},
  journal   = {Journal of Applied Remote Sensing},
  publisher = {International Society for Optics and Photonics},
  year      = {2016},
}

@Article{luigi2020paper3,
  author    = {Luppino, Luigi T and Hansen, Mads A and Kampffmeyer, Michael and Bianchi, Filippo M and Moser, Gabriele and Jenssen, Robert and Anfinsen, Stian N},
  title     = {Code-Aligned Autoencoders for Unsupervised Change Detection in Multimodal Remote Sensing Images},
  journal   = {IEEE Transactions on Neural Networks and Learning Systems},
  year      = {2022},
}

@Article{camps2008kernel,
  author    = {Camps-Valls, Gustavo and G{\'o}mez-Chova, Luis and Mu{\~n}oz-Mar{\'\i}, Jordi and Rojo-{\'A}lvarez, Jos{\'e} Luis and Mart{\'\i}nez-Ram{\'o}n, Manel},
  title     = {Kernel-based framework for multitemporal and multisource remote sensing data classification and change detection},
  number    = {6},
  pages     = {1822--1835},
  volume    = {46},
  journal   = {IEEE Trans.\ Geosci.\ Remote Sens.},
  publisher = {IEEE},
  year      = {2008},
}

@Article{coppin2004review,
  author    = {Coppin, Pol and Jonckheere, Inge and Nackaerts, Kristiaan and Muys, Bart and Lambin, Eric},
  title     = {Digital change detection methods in ecosystem monitoring: a review},
  number    = {9},
  pages     = {1565--1596},
  volume    = {25},
  journal   = {Int.\ J.\ Remote Sens.},
  publisher = {Taylor \& Francis},
  year      = {2004},
}

@Article{hall2016remote,
  author    = {Hall, RJ and Castilla, G and White, JC and Cooke, BJ and Skakun, RS},
  title     = {Remote sensing of forest pest damage: a review and lessons learned from a {C}anadian perspective},
  number    = {S1},
  pages     = {S296--S356},
  volume    = {148},
  journal   = {The Canadian Entomologist},
  publisher = {Cambridge University Press},
  year      = {2016},
}

@Article{ran2018kernel,
  author    = {Ran, Qiong and Li, Wei and Du, Qian},
  title     = {Kernel one-class weighted sparse representation classification for change detection},
  number    = {6},
  pages     = {597--606},
  volume    = {9},
  journal   = {Remote Sensing Letters},
  publisher = {Taylor \& Francis},
  year      = {2018},
}

@Article{jian2021gan,
  author    = {Jian, Ping and Chen, Keming and Cheng, Wei},
  title     = {{GAN}-Based One-Class Classification for Remote-Sensing Image Change Detection},
  journal   = {IEEE Geosci.\ Remote Sens.\ Lett.},
  publisher = {IEEE},
  year      = {2021},
}

@Article{munoz2010semisupervised,
  author    = {M{\~u}noz-Mar{\'\i}, Jordi and Bovolo, Francesca and G{\'o}mez-Chova, Luis and Bruzzone, Lorenzo and Camp-Valls, Gustavo},
  title     = {Semisupervised one-class support vector machines for classification of remote sensing data},
  number    = {8},
  pages     = {3188--3197},
  volume    = {48},
  journal   = {IEEE Trans.\ Geosci.\ Remote Sens.},
  publisher = {IEEE},
  year      = {2010},
}

@Article{mitchell2017current,
  author    = {Mitchell, Anthea L and Rosenqvist, Ake and Mora, Brice},
  title     = {Current remote sensing approaches to monitoring forest degradation in support of countries measurement, reporting and verification ({MRV}) systems for {REDD}+},
  number    = {1},
  pages     = {1--22},
  volume    = {12},
  journal   = {Carbon balance and management},
  publisher = {BioMed Central},
  year      = {2017},
}

@Book{Bishop2006,
  author    = {Bishop, Christopher},
  title     = {Pattern Recognition and Machine Learning},
  isbn      = {0387310738},
  publisher = {Springer},
  address   = {New York},
  year      = {2006},
}

@Article{jepsen2013ecosystem,
  author    = {Jepsen, Jane U and Biuw, Martin and Ims, Rolf A and Kapari, Lauri and Schott, Tino and Vindstad, Ole Petter L and Hagen, Snorre B},
  title     = {Ecosystem impacts of a range expanding forest defoliator at the forest-tundra ecotone},
  number    = {4},
  pages     = {561--575},
  volume    = {16},
  journal   = {Ecosystems},
  publisher = {Springer},
  year      = {2013},
}

@Article{senf2017remote,
  author    = {Senf, Cornelius and Seidl, Rupert and Hostert, Patrick},
  title     = {Remote sensing of forest insect disturbances: current state and future directions},
  pages     = {49--60},
  volume    = {60},
  journal   = {Int.\ J.\ Appl.\ Earth Observ.\ Geoinf.},
  publisher = {Elsevier},
  year      = {2017},
}

@Article{henden2020end,
  author    = {Henden, John-Andr{\'e} and Ims, Rolf A and Yoccoz, Nigel G and Asbj{\o}rnsen, Einar J and Stien, Audun and Mellard, Jarad Pope and Tveraa, Torkild and Marolla, Filippo and Jepsen, Jane Uhd},
  title     = {End-user involvement to improve predictions and management of populations with complex dynamics and multiple drivers},
  number    = {6},
  pages     = {e02120},
  volume    = {30},
  journal   = {Ecological Applications},
  publisher = {Wiley Online Library},
  year      = {2020},
}

@Article{li2020one,
  author    = {Li, Wenkai and Guo, Qinghua and Elkan, Charles},
  title     = {One-Class Remote Sensing Classification from Positive and Unlabeled Background Data},
  volume    = {14},
  journal   = {IEEE J.\ Sel.\ Topics in Appl.\ Earth Observ.\ Remote Sens.},
  publisher = {IEEE},
  year      = {2020},
}

@Article{jepsen2011rapid,
  Title                    = {Rapid northwards expansion of a forest insect pest attributed to spring phenology matching with sub-Arctic birch},
  Author                   = {Jepsen, Jane U and Kapari, Lauri and Hagen, Snorre B and Schott, Tino and Vindstad, Ole Petter L and Nilssen, Arne C and Ims, Rolf A},
  Journal                  = {Global Change Biology},
  Year                     = {2011},
  Number                   = {6},
  Pages                    = {2071--2083},
  Volume                   = {17},
  Publisher                = {Wiley Online Library}
}

@Article{jepsen2008moth,
  Title                    = {Climate change and outbreaks of the geometrids Operophtera brumata and Epirrita autumnata in subarctic birch forest: evidence of a recent outbreak range expansion},
  Author                   = {Jepsen, Jane Uhd and Hagen, Snorre B. and Ims, Rolf A. and Yoccoz, Nigel G.},
  Journal                  = {Journal of Animal Ecology},
  Year                     = {2008},
  Number                   = {2},
  Pages                    = {257--264},
  Volume                   = {77},

  Publisher                = {Wiley Online Library}
}

@Book{aba2013caff,
  author    = {CAFF},
  publisher = {Conservation of Arctic Flora and Fauna},
  title     = {Arctic Biodiversity Assessment 2013},
  year      = {2013},
}

\end{document}